\documentclass[11pt]{article}

\usepackage[]{acl}

\usepackage{algorithm}
\usepackage{subcaption}
\usepackage{graphicx}
\usepackage{amssymb}
\usepackage{amsthm}
\usepackage{hyperref}
\usepackage{booktabs}
\usepackage{url}
\usepackage{xcolor}
\usepackage[table]{xcolor}
\usepackage{listings}
\usepackage{float}
\usepackage{amsmath}
\usepackage{times}
\usepackage{latexsym}
\usepackage{tcolorbox}

\usepackage[T1]{fontenc}

\usepackage[utf8]{inputenc}
\usepackage{multirow}
\usepackage{microtype}

\usepackage{inconsolata}

\usepackage{graphicx}

\title{Functional Entropy: Predicting Functional Correctness in LLM-Generated Code with Uncertainty Quantification}

\author{Dylan Bouchard\thanks{Co-first authors}
\qquad
Mohit Singh Chauhan$^*$\thanks{Corresp. to \texttt{mohitsingh.chauhan@cvshealth.com}} \qquad
Zeya Ahmad \qquad
Ho-Kyeong Ra 
\\ 
CVS Health\textsuperscript{\textregistered}, Wellesley, MA, USA 
}

\begin{document}
\maketitle
\begin{abstract}
Large language models have shown impressive capabilities in code generation, yet they often produce functionally incorrect code. Uncertainty quantification (UQ) methods have emerged as a promising approach for detecting hallucinations in natural language generation, but their effectiveness for code generation tasks remains underexplored. We systematically evaluate how UQ techniques transfer to code generation across three programming languages, five LLMs, and over 1,700 problems. We find that some token-probability-based methods generalize effectively without modification, while sampling-based methods relying on natural language inference (NLI) fail because NLI models cannot distinguish functionally different code, causing most responses to collapse into a single semantic cluster.  To address this, we introduce \emph{functional equivalence methods}, a family of code-specific methods that replace NLI-based semantic equivalence with an LLM-based functional equivalence assessment, including functional entropy, a code-specific analog of semantic entropy. Functional equivalence methods achieve top AUROC in 11 out of 15 model-benchmark combinations and the best calibration across most settings, consistently outperforming both NLI-based counterparts and all other methods evaluated.
\end{abstract}


\begin{figure*}[t]
    \centering
    \includegraphics[width=0.9\textwidth]{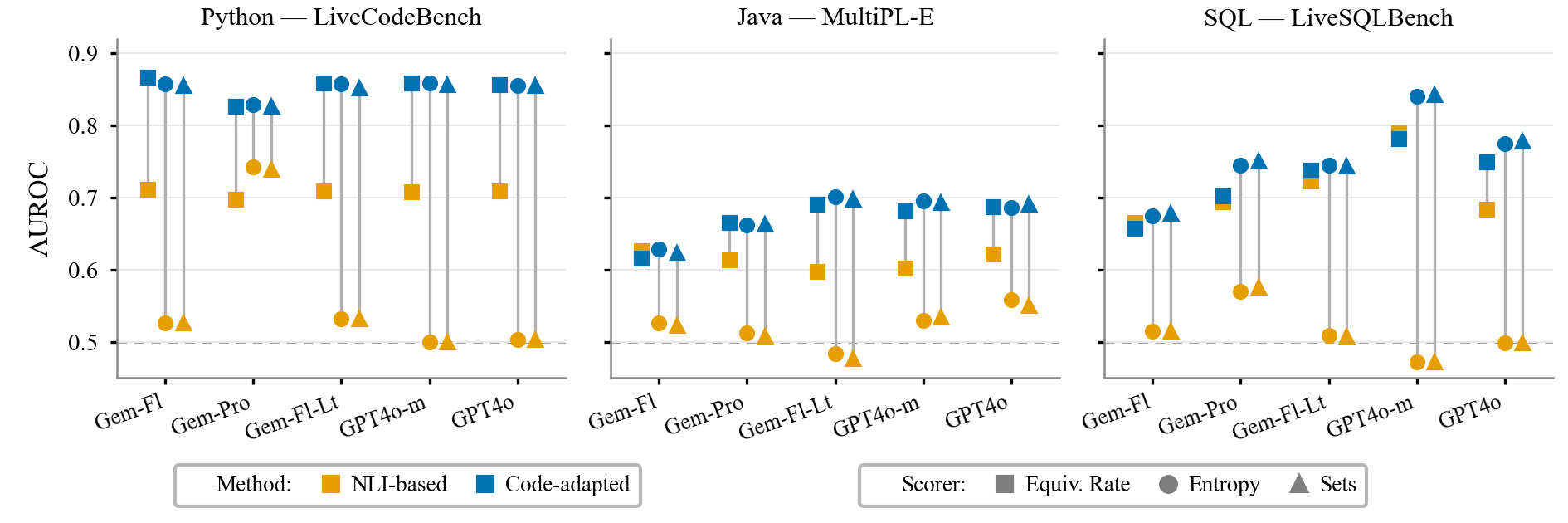}
    \caption{Code correctness AUROC of NLI-based vs. code-adapted methods. 
Equivalence rate measures agreement between the original and sampled responses; Entropy and Sets 
measure clustering-based consistency.}
    \label{fig:nlg_vs_func}
\end{figure*}

\section{Introduction}

Large language models (LLMs) have demonstrated remarkable capabilities in code generation, powering tools that assist developers in writing, completing, and debugging code. However, these models can produce plausible-looking but functionally incorrect code, a failure mode that poses significant risks in software development contexts where correctness is paramount. Studies on established benchmarks such as LiveCodeBench \citep{jain2024livecodebenchholisticcontaminationfree} and LiveSQLBench \citep{livesqlbench2025} show that even state-of-the-art models produce incorrect solutions for a substantial fraction of problems \citep{gao2025systematic, tian2025codehaluinvestigatingcodehallucinations}. Incorrect code can fail silently, introduce security vulnerabilities, or cause downstream system failures. These risks are amplified in settings where generated code is executed with limited or no human review, such as autonomous coding agents, text-to-SQL systems that run queries directly against databases, or agentic workflows where code is generated and executed as an intermediate tool-use step. Reliably predicting when generated code is likely wrong is therefore a pressing practical challenge.

In natural language generation (NLG), uncertainty quantification (UQ) has emerged as a promising approach for detecting unreliable outputs. Methods such as predictive entropy, self-consistency, and semantic entropy leverage token probabilities or sample agreement to estimate model confidence, with low confidence serving as a signal for potential hallucination. Recent work has shown that these techniques can effectively identify unreliable generations in question-answering, summarization, and other NLG tasks~\cite{Farquhar2024, Vashurin_2025}. Yet it remains unclear how well these methods transfer to code generation, where outputs have distinct structural properties and where functional correctness, as determined by test execution, provides a relatively unambiguous ground truth. While recent studies have begun exploring UQ for code generation, existing evaluations remain limited in scope, leaving practitioners without comprehensive guidance on which techniques to adopt across different languages and models or how to adapt them.

Several factors complicate this transfer. First, code exhibits \emph{functional equivalence}, meaning syntactically different programs can be semantically identical, and hence surface-level disagreement among samples does not necessarily indicate uncertainty. Second, UQ methods designed for natural language often rely on NLG-specific components, such as natural language inference (NLI) models to assess semantic equivalence or text embeddings to measure similarity. When applied to code without adaptation, these components become a bottleneck, degrading the performance of otherwise effective UQ frameworks. Third, although execution-based evaluation provides the most reliable measure of code correctness, gold test cases are not always available, and automatically generated tests may not cover the relevant failure modes. This raises the question of whether execution-free UQ methods can approximate the signal that test suites provide, and under what conditions they fall short.

These complications motivate not only a systematic evaluation of UQ techniques in the code domain, but also the development of code-specific adaptations where general-purpose methods prove insufficient. In this paper, we investigate how existing uncertainty quantification techniques generalize to code generation across three programming languages (Python, Java, and SQL). We evaluate a comprehensive suite of methods spanning token-probability-based approaches, sampling-based consistency methods, and reflexive techniques across five LLMs and established code generation benchmarks. We find that some single-generation token-probability-based methods transfer effectively to code with no modification, though their effectiveness varies across settings. For sampling-based methods, we show that NLI-based semantic equivalence is the critical bottleneck because NLI models cannot reliably distinguish functionally different code, causing most responses to collapse into a single cluster and rendering clustering-based UQ methods uninformative. To address this, we introduce \emph{functional equivalence methods}, a family of code-specific methods that replace NLI-based semantic equivalence with LLM-based functional equivalence assessment, including functional entropy, a code-specific analog of semantic entropy. Functional equivalence methods consistently achieve the strongest performance across models and languages, demonstrating that the bottleneck in transferring sampling-based UQ methods to code lies in the equivalence assessment mechanism.

Our contributions are as follows. First, we provide a comprehensive evaluation of uncertainty quantification methods for code correctness prediction across five LLMs, three programming languages, and over 1,700 problems, establishing baseline performance and revealing that method effectiveness varies substantially across languages, offering actionable guidance for practitioners. Second, we introduce functional equivalence methods, a family of methods comprising functional equivalence rate, functional entropy, and functional sets confidence, that replace NLI-based semantic equivalence with LLM-evaluated functional equivalence for clustering code responses. These methods achieve top AUROC in 11 out of 15 model-benchmark combinations and the best calibration across most settings, consistently outperforming both their NLI-based counterparts and all other methods evaluated. Third, we release all methods, including functional entropy and additional code-adapted scorers using code embeddings and CodeBLEU-based consistency \citep{sharma2025assessing}, as 
part of UQLM \citep{bouchard2026uqlmpythonpackageuncertainty}, 
our open-source library for uncertainty quantification in both natural language and code generation.\footnote{
\url{https://github.com/cvs-health/uqlm}}

\section{Related Work}
\paragraph{Uncertainty Quantification for Natural Language Generation.}

Uncertainty quantification has demonstrated strong performance in detecting hallucinations in natural language generation \citep{huang2023surveyhallucinationlargelanguage, shorinwa2024surveyuncertaintyquantificationlarge}. Existing methods can be broadly categorized by their access requirements and computational approach. Various white-box methods leverage token probabilities from a single generation, using signals such as sequence probability \citep{malinin2021uncertaintyestimationautoregressivestructured}, minimum token probability \citep{manakul2023selfcheckgptzeroresourceblackboxhallucination}, probability margin \citep{farr2024redctsystemsdesignmethodology}, and token-level entropy \citep{scalena2025eagerentropyawaregenerationadaptive}. Sampling-based methods generate multiple responses from the same prompt and measure consistency, either through black-box approaches using exact match \citep{cole2023selectivelyansweringambiguousquestionsy}, embedding similarity \citep{manakul2023selfcheckgptzeroresourceblackboxhallucination, bouchard2025uncertaintyquantificationlanguagemodels}, or NLI-based semantic equivalence \citep{chen2023quantifyinguncertaintyanswerslanguage, kuhn2023semanticuncertaintylinguisticinvariances, Farquhar2024}, or through white-box approaches that additionally incorporate token probabilities \citep{Vashurin_2025, vashurin2025uncertaintyquantificationllmsminimum, qiu2024semanticdensityuncertaintyquantification}. Others have adapted these techniques for claim-level scoring of long-form generations \citep{jiang2024graphbaseduncertaintymetricslongform, bouchard2026finegraineduncertaintyquantificationlongform, zhang2024luqlongtextuncertaintyquantification, zhang2025atomiccalibrationllmslongform}. Reflexive methods prompt models to self-evaluate their outputs, either by eliciting explicit confidence scores \citep{tian2023justaskcalibrationstrategies} or by probing agreement with the original response \citep{kadavath2022languagemodelsmostlyknow}. While these methods have been extensively studied for text-generation, their effectiveness for code generation remains underexplored.

\paragraph{Hallucination Detection in Code Generation.}
Recent work has begun characterizing hallucinations in code generation, proposing taxonomies and benchmarks for evaluation \citep{gao2025systematic, tian2025codehaluinvestigatingcodehallucinations, liu2026functionalcorrectnessexploringhallucinations, jiang2024collubenchbenchmarkpredictinglanguage}. From an uncertainty perspective, \citet{Huang_2025} conducted an exploratory study of uncertainty estimation for LLMs across both NLP tasks and code generation, finding that uncertainty metrics can help identify buggy programs, while \citet{spiess2024calibrationcorrectnesslanguagemodels} find token-probability-based confidence scores for generated code are better-calibrated than those based on verbalized confidence. \citet{sharma2025assessing} proposed symbolic clustering for uncertainty estimation in code, arguing that existing confidence measures fail to capture semantic equivalence. 

Several approaches have been proposed to mitigate code hallucinations, including entropy-aware adaptive decoding \citep{scalena2025eagerentropyawaregenerationadaptive}, attention-based uncertainty quantification \citep{vazhentsev2025uncertainty}, and highlighting likely-incorrect tokens \citep{vasconcelos2025generation}. In adjacent domains, \citet{yang2025hallucinationdetectionllmbasedtexttosql} proposed metamorphic testing for hallucination detection in text-to-SQL generation, \citet{tantithamthavorn2026hallujudgereferencefreehallucinationdetection} developed a reference-free hallucination detector for code review automation, and \citet{healy2026internalrepresentationsindicatorshallucinations} explored internal-representation-based UQ for agent tool selection. From a decoding perspective, \citet{he2025adadec} proposed uncertainty-aware adaptive decoding, and \citet{zhu2025uncertainty} presented UnCerT-CoT for dynamic chain-of-thought reasoning. Our work differs by systematically evaluating a comprehensive suite of UQ methods across multiple programming languages and models, diagnosing the specific failure of NLI-based equivalence for code, and introducing functional entropy methods that recover strong performance through LLM-based functional equivalence assessment.

\section{Methods}
We consider the problem of predicting functional correctness in code generation (i.e., whether generated code satisfies the intended specification), often termed hallucination detection in the broader literature. Given a prompt $x$ (e.g., a natural language description or docstring) and a generated code response $y$, our goal is to produce a confidence score $s(y) \in [0,1]$ that effectively distinguishes functionally correct code from incorrect code, enabling reliable ranking and classification without requiring execution or access to test cases at inference time. Our primary evaluation criterion is AUROC, measuring how well the score ranks correct generations above incorrect ones.

We evaluate three families of uncertainty quantification methods: (1) token-probability-based methods that operate on a single generation, (2) sampling-based consistency methods that compare multiple generations from the same prompt, and (3) reflexive methods that prompt the model to self-evaluate its outputs. For sampling-based methods, we develop code-specific extensions that replace natural language components with analogs designed for code.

\subsection{Token-Probability-Based Methods}

These methods derive confidence scores directly from the token-level probabilities produced by the language model during generation. Let the tokenization of response $y$ be denoted $\{t_1, \ldots, t_L\}$, where $L$ is the number of tokens, and let $p_j$ denote the probability assigned to token $t_j$ by the model.

\paragraph{Length-Normalized Sequence Probability (LNSP).} Sequence probability \citep{Vashurin_2025} is the joint probability of all tokens in the response: 
\begin{equation*}
    \text{SP}(y; x) = \prod_{j=1}^{L} p_j.
\end{equation*}
 To account for the tendency of longer sequences to have lower joint probabilities, we focus on length-normalized sequence probability \citep{malinin2021uncertaintyestimationautoregressivestructured}, which computes the geometric mean of token probabilities:
\begin{equation*}
    \text{LNSP}(y; x) = \left( \prod_{j=1}^{L} p_j \right)^{1/L}
\end{equation*}

\paragraph{Minimum Token Probability (MTP).} This method uses the minimum token probability as a confidence score \citep{manakul2023selfcheckgptzeroresourceblackboxhallucination}, capturing the intuition that a single low-confidence token may indicate an unreliable generation:
\begin{equation*}
    \text{MTP}(y; x) = \min_{j \in \{1, \ldots, L\}} p_j
\end{equation*}

Other methods require access to the top-$K$ token probabilities for each token position. Let $\{p_{j,1}, \ldots, p_{j,K}\}$ denote the top-$K$ token probabilities at position $j$, ordered by decreasing probability.

\paragraph{Probability Margin (PM).} Probability margin \citep{farr2024redctsystemsdesignmethodology} measures the average gap between the most likely and second most likely token at each position:
\begin{equation*}
    \text{PM}(y; x) = \frac{1}{L} \sum_{j=1}^{L} (p_{j,1} - p_{j,2})
\end{equation*}

\paragraph{Average Token Negentropy at $K$ (ATN@$K$).} This method computes the entropy over the top-$K$ tokens at each position and normalizes it to a confidence score \citep{scalena2025eagerentropyawaregenerationadaptive, manakul2023selfcheckgptzeroresourceblackboxhallucination}. The top-$K$ token entropy at position $j$ is:
\begin{equation*}
    \text{TE@}K(t_j) = -\sum_{k=1}^{K} p_{j,k}^{(K)} \log p_{j,k}^{(K)}
\end{equation*}
where $p_{j,k}^{(K)}=p_{j,k} /\sum_{i=1}^K p_{j,i},$ $k=1,...,K$ are the renormalized top-K probabilities. The token negentropy transformation normalizes this to $[0,1]$ with higher values indicating higher confidence:
\begin{equation*}
    \text{TN@}K(t_j) = 1 - \frac{\text{TE@}K(t_j)}{\log K}
\end{equation*}
Average token negentropy at $K$ is then the mean top-$K$ token negentropy across all positions:
\begin{equation*}
    \text{ATN@}K(y; x) = \frac{1}{L} \sum_{j=1}^{L} \text{TN@}K(t_j)
\end{equation*}

\paragraph{Minimum Token Negentropy at $K$ (MTN@$K$).} Analogous to minimum token probability, this method uses the minimum token negentropy as the confidence score \citep{scalena2025eagerentropyawaregenerationadaptive, manakul2023selfcheckgptzeroresourceblackboxhallucination}:
\begin{equation*}
    \text{MTN@}K(y; x) = \min_{j \in \{1, \ldots, L\}} \text{TN@}K(t_j)
\end{equation*}

\subsection{Sampling-Based Consistency Methods}
\label{sec:consistency}

These methods generate $m$ additional responses $\tilde{\mathbf{y}} = \{\tilde{y}_1, \ldots, \tilde{y}_m\}$ from the same prompt $x$ using a non-zero sampling temperature (specified in Section~\ref{sec:experiments}), then measure consistency between the original response $y$ and the sampled responses. The intuition is that when a model is confident, independently sampled responses should agree; disagreement signals uncertainty.

\subsubsection{Functional Equivalence Methods}

We introduce a family of three methods built on a shared foundation: LLM-based functional equivalence assessment. In natural language UQ, sampling-based methods typically use NLI models to assess semantic equivalence between responses, classifying each pair as entailment, contradiction, or neutral \citep{kuhn2023semanticuncertaintylinguisticinvariances}. For code, we replace this with a direct LLM-based assessment of whether two code snippets are functionally equivalent, i.e., whether they would produce the same outputs for all valid inputs (see Appendix~\ref{sec:prompts} for the prompt template). Each method extracts a different summary statistic from the resulting equivalence structure.

\paragraph{Functional Equivalence Rate (FER).} This method estimates the proportion of sampled responses that are functionally equivalent to the original response:
\begin{equation*}
    \text{FER}(y; \tilde{\mathbf{y}}, x) = \frac{1}{m} \sum_{j=1}^{m} \mathbb{I}[y \equiv \tilde{y}_j]
\end{equation*}
where $\mathbb{I}[y \equiv \tilde{y}_j] \in \{0, 1\}$ denotes the LLM-based functional equivalence indicator.

\paragraph{Functional Entropy (FE).} Functional entropy clusters responses based on functional equivalence, then computes entropy over the cluster distribution.\footnote{
Clusters are formed through a greedy procedure, where the original response initializes the first cluster, and each remaining response is compared against existing cluster representatives. Responses judged equivalent to a representative join that cluster; the first response not equivalent to any existing representative initializes a new cluster and becomes its representative. This process continues until all responses are assigned, requiring fewer comparisons than exhaustive pairwise evaluation.} This is a code-specific analogue of semantic entropy \citep{kuhn2023semanticuncertaintylinguisticinvariances}, replacing NLI-based mutual entailment with LLM-based functional equivalence assessment. Let $\mathcal{C}$ denote the set of clusters obtained from $(y, \tilde{y}_1, \ldots, \tilde{y}_m)$, and let $P(C)$ denote the proportion of responses in cluster $C$. Functional entropy is:
\begin{equation*}
    \text{FE}(y; \tilde{\mathbf{y}}, x) = -\sum_{C \in \mathcal{C}} P(C) \log P(C)
\end{equation*}
Following \citet{bouchard2025uncertaintyquantificationlanguagemodels}, we normalize to obtain a confidence score in $[0,1]$:
\begin{equation*}
    \text{NFN}(y; \tilde{\mathbf{y}}, x) = 1 - \frac{\text{FE}(y; \tilde{\mathbf{y}}, x)}{\log(m+1)}
\end{equation*}

\paragraph{Functional Sets Confidence (FSC).} This method counts the number of unique functional clusters $|\mathcal{C}|$ \citep{lin2024generatingconfidenceuncertaintyquantification}. Following \citet{bouchard2026uqlmpythonpackageuncertainty}, we normalize to $[0,1]$:
\begin{equation*}
    \text{FSC}(y; \tilde{\mathbf{y}}, x) = \frac{m + 1 - |\mathcal{C}|}{m}
\end{equation*}
When all $m+1$ responses (the original plus $m$ samples) cluster together ($|\mathcal{C}| = 1$), confidence is 1; when all responses are distinct ($|\mathcal{C}| = m + 1$), confidence is 0.

\subsubsection{Similarity-Based Methods}

\paragraph{Normalized Cosine Similarity (NCS).} This method embeds responses using a code embedding model $V: \mathcal{Y} \rightarrow \mathbb{R}^d$ and measures average cosine similarity across sampled responses, normalized to $[0,1]$ \citep{bouchard2025uncertaintyquantificationlanguagemodels}:
\begin{equation*}
    \text{NCS}(y; \tilde{\mathbf{y}}, x) = \frac{1}{2} + \frac{1}{2m} \sum_{j=1}^{m} \frac{V(y) \cdot V(\tilde{y}_j)}{\|V(y)\| \cdot \|V(\tilde{y}_j)\|}.
\end{equation*}

\paragraph{CodeBLEU Consistency (CBC).} Following \citet{sharma2025assessing}, we implement a consistency scorer based on CodeBLEU \citep{ren2020codebleumethodautomaticevaluation}, a metric for code generation that incorporates n-gram match, weighted n-gram match, syntax match via abstract syntax trees, and semantic match via data-flow analysis. Unlike the equivalence-based methods above, which assess functional agreement, CBC captures structural and syntactic similarity between generations:
\begin{equation*}
    \text{CBC}(y; \tilde{\mathbf{y}}, x) = \frac{1}{m} \sum_{j=1}^{m} \text{CodeBLEU}(y, \tilde{y}_j)
\end{equation*}

\subsection{Reflexive Methods}
Reflexive methods prompt the model to evaluate the correctness of its own outputs. 

\paragraph{P(True).} This method asks the model to classify its response as correct or incorrect, using the token probability assigned to ``True'' as the confidence score: $\text{P(True)}(y; x) = p(\text{``True''} \mid x, y)$ \citep{kadavath2022languagemodelsmostlyknow}.

\paragraph{Verbalized Confidence (VC).} VC prompts the model to express confidence on a six-level Likert scale from ``No chance'' (0\%) to ``Almost certain'' (100\%), mapped to $[0,1]$ without requiring access to token-level probabilities \citep{tian2023justaskcalibrationstrategies}. 




\section{Experiments}
\label{sec:experiments}
We evaluate the full suite of uncertainty quantification methods on code generation tasks spanning three programming languages, five large language models, and 1,711 problems.

\begin{table}[t]
\tiny
\centering
\begin{tabular}{lccccc}
\toprule
Benchmark & Gem-Pro & Gem-Fl & Gem-Fl-Lt & GPT4o & GPT4o-m \\
\midrule
LCB (Python) & 0.91 & 0.87 & 0.67 & 0.56 & 0.51   \\
Mult.-E (Java)      & 0.53 & 0.51 & 0.48 & 0.57 & 0.56 \\
LSQLB (SQL)    & 0.45 & 0.45  & 0.24 & 0.29 & 0.21 \\
\bottomrule
\end{tabular}
\caption{Model accuracy (Pass@1) across benchmarks.}
\label{tab:accuracy}
\end{table}

\subsection{Experimental Setup}

We evaluate on three code generation benchmarks covering Python, Java, and SQL. \textbf{LiveCodeBench} \citep{jain2024livecodebenchholisticcontaminationfree} contains 1,055 Python problems sourced from LeetCode, AtCoder, and Codeforces programming platforms. Problems span a range of difficulty levels and require generating complete function implementations from natural language descriptions and function signatures. \textbf{MultiPL-E (Java)} \citep{cassano2022multiplescalableextensibleapproach} provides 386 Java problems translated from the Mostly Basic Python Problems benchmark, covering common programming tasks such as string manipulation, list processing, and mathematical computations. \textbf{LiveSQLBench} \citep{livesqlbench2025} contains 270 SQLite problems requiring database query generation from natural language questions and schema descriptions; problems range from simple single-table queries to complex multi-table joins and aggregations.

We generate code using five large language models spanning different capability levels and providers: GPT-4o, GPT-4o-mini \citep{OpenAI}, Gemini-2.5-Pro, Gemini-2.5-Flash, and Gemini-2.5-Flash-Lite \citep{gemini_doc}. For each model-benchmark pair, we generate an original response using greedy decoding and $m=10$ sampled responses with temperature $T=1.0$ for sampling-based methods. We ablate over $m \in \{1, \ldots, 10\}$ in Appendix~\ref{sec:ablations}. For Python and Java benchmarks, a generated solution is considered correct if it passes all unit tests for the given problem. For SQL problems, correctness is determined by comparing the output of the generated query against that of the reference solution.

We evaluate each scorer on classification performance and calibration. For classification, we report Area Under the Receiver Operating Characteristic curve (AUROC), treating correctness prediction as a binary classification task. For calibration, we report Brier Score and Expected Calibration Error (ECE), which measure how well confidence scores align with empirical accuracy. Calibration results are reported in Appendix~\ref{sec:calibration}. We use Gemini-2.5-Flash for the functional equivalence assessments, validated against execution-based outcomes with 93--97\% precision across evaluated settings (Appendix~\ref{sec:validation}), and \texttt{jina-embeddings-v2-base-code} for code embeddings. 

Across all experiments, functional equivalence methods achieve top AUROC in 11 out of 15 model-benchmark combinations. These methods also achieve the best calibration in Python and SQL while remaining competitive on Java, meaning their confidence scores often serve as reliable probability estimates in addition to providing strong ranking performance. We now present results for each benchmark; cross-benchmark patterns and implications are discussed in Section~\ref{sec:discussion}.

\subsection{Python (LiveCodeBench)}

\begin{table}[htp]
\tiny
\centering
\begin{tabular}{lrrrrr}
\toprule
Scorer & Gem-Fl & Gem-Pro & Gem-Fl-Lt & GPT4o-m & GPT4o \\
\midrule
\multicolumn{6}{l}{\textit{Token-Probability Methods}} \\
Seq. Prob. & \cellcolor{red!35!white}0.691 & \cellcolor{red!45!white}0.580 & \cellcolor{red!15!white}0.761 & \cellcolor{red!15!white}0.757 & \cellcolor{red!25!white}0.729 \\
Min. Prob. & \cellcolor{blue!5!white}0.773 & \cellcolor{red!15!white}0.762 & \cellcolor{blue!5!white}0.783 & \cellcolor{red!25!white}0.738 & \cellcolor{red!35!white}0.705 \\
Prob. Margin & \cellcolor{red!45!white}0.668 & \cellcolor{red!45!white}0.554 & \cellcolor{red!5!white}0.762 & \cellcolor{red!25!white}0.752 & \cellcolor{red!25!white}0.749 \\
Avg. Tok. Neg. & \cellcolor{red!35!white}0.695 & \cellcolor{red!45!white}0.582 & \cellcolor{blue!5!white}0.774 & \cellcolor{blue!5!white}0.776 & \cellcolor{red!15!white}0.760 \\
Min. Tok. Neg. & \cellcolor{red!5!white}0.766 & \cellcolor{red!5!white}0.770 & \cellcolor{blue!15!white}0.797 & \cellcolor{blue!5!white}0.779 & \cellcolor{red!5!white}0.768 \\
\midrule
\multicolumn{6}{l}{\textit{Functional Equivalence Methods}} \\
Func. Equiv. & \cellcolor{blue!45!white}\underline{0.867} & \cellcolor{blue!25!white}0.826 & \cellcolor{blue!45!white}\underline{0.859} & \cellcolor{blue!45!white}0.859 & \cellcolor{blue!35!white}\underline{0.856} \\
Func. Entropy & \cellcolor{blue!45!white}0.857 & \cellcolor{blue!35!white}\underline{0.829} & \cellcolor{blue!45!white}0.857 & \cellcolor{blue!45!white}\underline{0.859} & \cellcolor{blue!35!white}0.855 \\
Func. Sets & \cellcolor{blue!35!white}0.855 & \cellcolor{blue!25!white}0.826 & \cellcolor{blue!35!white}0.852 & \cellcolor{blue!45!white}0.856 & \cellcolor{blue!35!white}0.855 \\
\midrule
\multicolumn{6}{l}{\textit{Similarity-Based Methods}} \\
Cos. Sim. & \cellcolor{red!15!white}0.753 & \cellcolor{blue!15!white}0.804 & \cellcolor{red!15!white}0.756 & \cellcolor{red!25!white}0.738 & \cellcolor{red!35!white}0.724 \\
CodeBLEU & \cellcolor{blue!15!white}0.795 & \cellcolor{blue!25!white}0.813 & \cellcolor{blue!15!white}0.795 & \cellcolor{red!5!white}0.766 & \cellcolor{red!25!white}0.744 \\
\midrule
\multicolumn{6}{l}{\textit{Reflexive Methods}} \\
Verb. Conf. & \cellcolor{blue!15!white}0.795 & \cellcolor{red!35!white}0.716 & \cellcolor{red!45!white}0.597 & \cellcolor{red!35!white}0.725 & \cellcolor{red!45!white}0.681 \\
P(True) & \cellcolor{red!5!white}0.771 & \cellcolor{blue!25!white}0.815 & \cellcolor{blue!35!white}0.830 & \cellcolor{red!15!white}0.752 & \cellcolor{blue!25!white}0.824 \\
\bottomrule
\end{tabular}
\caption{Code correctness AUROC on LiveCodeBench (Python). Higher is better; red = low, blue = high.}
\label{tab:auroc_python}
\end{table}

Table~\ref{tab:auroc_python} presents AUROC results for predicting functional correctness on LiveCodeBench. Model accuracy (pass@1) ranges from 0.51 (GPT-4o-mini) to 0.91 (Gemini-2.5-Pro). Overall AUROC scores are the highest across our three benchmarks, with most methods exceeding 0.70 and the best performers reaching 0.87, suggesting that the richness and diversity of Python solutions provides strong signal for uncertainty estimation.

Sampling-based methods using LLM-based functional equivalence achieve the strongest performance across all models. Functional equivalence rate achieves the highest overall score of 0.87 on Gemini-2.5-Flash, with functional entropy and functional sets confidence performing comparably (0.83--0.86 across models). These methods show remarkable consistency, maintaining strong performance regardless of underlying model capability. CodeBLEU consistency also performs well, achieving 0.81 on Gemini-2.5-Pro and 0.80 on Gemini-2.5-Flash, confirming that syntax-aware similarity captures meaningful variation in Python code.

Most token-probability methods show solid performance on Python. Minimum token negentropy achieves 0.77--0.80 across models, while minimum probability reaches 0.78 on Gemini-2.5-Flash-Lite. Sequence probability and probability margin are notably weaker on Gemini-2.5-Pro (0.580 and 0.554), indicating that aggregated token-level confidence does not reliably reflect correctness for all models. Among reflexive methods, P(True) is competitive, reaching 0.83 on Gemini-2.5-Flash-Lite and 0.82 on GPT-4o, while verbalized confidence is inconsistent (0.60--0.80).

\subsection{Java (MultiPL-E)}
\begin{table}[htp]
\tiny
\centering
\begin{tabular}{lrrrrr}
\toprule
Scorer & Gem-Fl & Gem-Pro & Gem-Fl-Lt & GPT4o-m & GPT4o \\
\midrule
\multicolumn{6}{l}{\textit{Token-Probability Methods}} \\
Seq. Prob. & \cellcolor{red!45!white}0.551 & \cellcolor{red!35!white}0.588 & \cellcolor{blue!5!white}0.637 & \cellcolor{red!35!white}0.572 & \cellcolor{blue!5!white}0.638 \\
Min. Prob. & \cellcolor{blue!5!white}0.635 & \cellcolor{red!15!white}0.618 & \cellcolor{blue!45!white}\underline{0.705} & \cellcolor{red!25!white}0.599 & \cellcolor{blue!15!white}0.647 \\
Prob. Margin & \cellcolor{red!45!white}0.525 & \cellcolor{red!25!white}0.596 & \cellcolor{red!25!white}0.601 & \cellcolor{red!35!white}0.553 & \cellcolor{blue!5!white}0.640 \\
Avg. Tok. Neg. & \cellcolor{red!45!white}0.536 & \cellcolor{red!25!white}0.598 & \cellcolor{red!25!white}0.589 & \cellcolor{red!45!white}0.551 & \cellcolor{blue!15!white}0.644 \\
Min. Tok. Neg. & \cellcolor{blue!15!white}0.652 & \cellcolor{blue!5!white}0.644 & \cellcolor{blue!25!white}0.674 & \cellcolor{red!5!white}0.632 & \cellcolor{blue!15!white}0.653 \\
\midrule
\multicolumn{6}{l}{\textit{Functional Equivalence Methods}} \\
Func. Equiv. & \cellcolor{red!15!white}0.616 & \cellcolor{blue!25!white}\underline{0.665} & \cellcolor{blue!35!white}0.690 & \cellcolor{blue!35!white}0.681 & \cellcolor{blue!35!white}0.687 \\
Func. Entropy & \cellcolor{red!5!white}0.628 & \cellcolor{blue!25!white}0.662 & \cellcolor{blue!45!white}0.701 & \cellcolor{blue!45!white}\underline{0.695} & \cellcolor{blue!35!white}0.686 \\
Func. Sets & \cellcolor{red!5!white}0.623 & \cellcolor{blue!25!white}0.663 & \cellcolor{blue!45!white}0.697 & \cellcolor{blue!45!white}0.693 & \cellcolor{blue!35!white}\underline{0.691} \\
\midrule
\multicolumn{6}{l}{\textit{Similarity-Based Methods}} \\
Cos. Sim. & \cellcolor{blue!15!white}0.645 & \cellcolor{red!5!white}0.627 & \cellcolor{red!15!white}0.615 & \cellcolor{red!15!white}0.610 & \cellcolor{red!5!white}0.626 \\
CodeBLEU & \cellcolor{blue!35!white}\underline{0.690} & \cellcolor{blue!25!white}0.662 & \cellcolor{blue!25!white}0.676 & \cellcolor{blue!15!white}0.658 & \cellcolor{blue!25!white}0.672 \\
\midrule
\multicolumn{6}{l}{\textit{Reflexive Methods}} \\
Verb. Conf. & \cellcolor{red!5!white}0.623 & \cellcolor{red!45!white}0.548 & \cellcolor{red!25!white}0.595 & \cellcolor{red!15!white}0.615 & \cellcolor{red!25!white}0.592 \\
P(True) & \cellcolor{red!45!white}0.529 & \cellcolor{red!15!white}0.616 & \cellcolor{red!45!white}0.548 & \cellcolor{blue!5!white}0.640 & \cellcolor{blue!35!white}0.687 \\
\bottomrule
\end{tabular}
\caption{Code correctness AUROC on MultiPL-E (Java). Higher is better; red = low, blue = high.}
\label{tab:auroc_java}
\end{table}

Table~\ref{tab:auroc_java} presents AUROC results for predicting functional correctness on MultiPL-E (Java). Model accuracy ranges from 0.49 (Gemini-2.5-Flash-Lite) to 0.57 (GPT-4o). Overall AUROC scores are lower than on Python problems, with most methods falling between 0.55--0.70 and limited separation between scorer families. As with Python, sampling-based methods using LLM-based functional equivalence achieve the strongest performance, though the margin over token-probability methods is less pronounced. Functional entropy and functional sets confidence reach approximately 0.70 on Gemini-2.5-Flash-Lite and GPT-4o-mini, while functional equivalence rate performs comparably. CodeBLEU consistency shows competitive results, achieving 0.69 on Gemini-2.5-Flash, suggesting that syntax-aware similarity metrics capture meaningful variation in Java code.

Token-probability methods perform moderately well on Java, with minimum probability achieving the highest AUROC of 0.705 on Gemini-2.5-Flash-Lite, outperforming several sampling-based methods for that model. Minimum token negentropy is consistent across models, ranging from 0.63 to 0.67. Among reflexive methods, P(True) achieves its best performance on GPT-4o (0.687), with ranking of AUROC across models mostly consistent with pass@1 rates, while verbalized confidence is consistently weaker (0.55--0.62).

\subsection{SQL (LiveSQLBench)}
\begin{table}[t]
\tiny
\centering
\begin{tabular}{lrrrrr}
\toprule
Scorer & Gem-Fl & Gem-Pro & Gem-Fl-Lt & GPT4o-m & GPT4o \\
\midrule
\multicolumn{6}{l}{\textit{Token-Probability Methods}} \\
Seq. Prob. & \cellcolor{red!45!white}0.554 & \cellcolor{red!45!white}0.529 & \cellcolor{red!15!white}0.658 & \cellcolor{blue!5!white}0.678 & \cellcolor{red!25!white}0.646 \\
Min. Prob. & \cellcolor{red!25!white}0.628 & \cellcolor{red!35!white}0.602 & \cellcolor{red!25!white}0.641 & \cellcolor{red!5!white}0.672 & \cellcolor{blue!5!white}0.679 \\
Prob. Margin & \cellcolor{red!45!white}0.546 & \cellcolor{red!45!white}0.517 & \cellcolor{red!25!white}0.629 & \cellcolor{red!5!white}0.673 & \cellcolor{red!35!white}0.616 \\
Avg. Tok. Neg. & \cellcolor{red!45!white}0.558 & \cellcolor{red!45!white}0.528 & \cellcolor{red!15!white}0.662 & \cellcolor{blue!5!white}0.694 & \cellcolor{red!25!white}0.632 \\
Min. Tok. Neg. & \cellcolor{red!15!white}0.654 & \cellcolor{red!35!white}0.623 & \cellcolor{blue!15!white}0.712 & \cellcolor{blue!5!white}0.687 & \cellcolor{blue!15!white}0.706 \\
\midrule
\multicolumn{6}{l}{\textit{Functional Equivalence Methods}} \\
Func. Equiv. & \cellcolor{red!15!white}0.657 & \cellcolor{blue!15!white}0.702 & \cellcolor{blue!25!white}0.738 & \cellcolor{blue!45!white}0.782 & \cellcolor{blue!35!white}0.749 \\
Func. Entropy & \cellcolor{red!5!white}0.674 & \cellcolor{blue!35!white}0.745 & \cellcolor{blue!35!white}0.745 & \cellcolor{blue!45!white}0.840 & \cellcolor{blue!45!white}0.775 \\
Func. Sets & \cellcolor{red!5!white}\underline{0.678} & \cellcolor{blue!35!white}\underline{0.750} & \cellcolor{blue!25!white}0.743 & \cellcolor{blue!45!white}\underline{0.842} & \cellcolor{blue!45!white}0.778 \\
\midrule
\multicolumn{6}{l}{\textit{Similarity-Based Methods}} \\
Cos. Sim. & \cellcolor{red!15!white}0.665 & \cellcolor{blue!25!white}0.726 & \cellcolor{blue!5!white}0.683 & \cellcolor{blue!15!white}0.705 & \cellcolor{blue!25!white}0.721 \\
\midrule
\multicolumn{6}{l}{\textit{Reflexive Methods}} \\
Verb. Conf. & \cellcolor{red!45!white}0.535 & \cellcolor{red!35!white}0.596 & \cellcolor{blue!25!white}0.739 & \cellcolor{blue!25!white}0.737 & \cellcolor{red!35!white}0.596 \\
P(True) & \cellcolor{red!15!white}0.647 & \cellcolor{blue!15!white}0.695 & \cellcolor{blue!35!white}\underline{0.749} & \cellcolor{blue!5!white}0.687 & \cellcolor{blue!45!white}\underline{0.818} \\
\bottomrule
\end{tabular}
\caption{Code correctness AUROC on LiveSQLBench (SQLite). Higher is better; red = low, blue = high.}
\label{tab:auroc_sql}
\end{table}

Table~\ref{tab:auroc_sql} presents AUROC results for classifying functional correctness on LiveSQLBench. Model accuracy on this benchmark ranges from 0.21 (GPT-4o-mini) to 0.45 (Gemini-2.5-Pro and Gemini-2.5-Flash). Overall AUROC scores are moderate, with most methods falling in the 0.55--0.75 range and the best performers reaching 0.84. Again, sampling-based methods using LLM-based equivalence assessment consistently achieve the highest scores across all models, while token-probability methods show weaker discrimination for SQL generation compared to Python and Java.

Among sampling-based methods, functional sets confidence and functional entropy achieve the strongest performance. Notably, GPT-4o-mini reaches 0.842 AUROC with functional sets confidence despite having the lowest accuracy (0.21); we discuss the practical implications of this finding in Section~\ref{sec:discussion}. Functional equivalence rate and cosine similarity also perform well, with scores above 0.70 for most models.\footnote{We omit CodeBLEU, as it does not support SQL.}

Token-probability methods show limited effectiveness for SQL, with most scores in the 0.52--0.71 range. Sequence probability and probability margin perform near chance level for Gemini-2.5-Pro and Gemini-2.5-Flash (0.52--0.55), suggesting that token-level confidence scores carry little discriminative signal for SQL correctness on these models. Minimum token negentropy is the strongest performer among single-generation methods, achieving 0.71 on Gemini-2.5-Flash-Lite and GPT-4o. Among reflexive methods, P(True) achieves strong performance on GPT-4o (0.82), outperforming most other methods for that model, while verbalized confidence shows inconsistent results, ranging from 0.54 on Gemini-2.5-Flash to 0.74 on Gemini-2.5-Flash-Lite.

\subsection{NLI Methods vs.\ Code-Adapted Methods}

To validate that replacing NLI-specific components with code-adapted analogs yields meaningful gains, we compare each functional equivalence method against its NLI-based counterpart: functional equivalence rate vs.\ non-contradiction probability \citep{chen2023quantifyinguncertaintyanswerslanguage}, functional entropy vs.\ NLI-based semantic entropy \citep{kuhn2023semanticuncertaintylinguisticinvariances}, and functional sets confidence vs.\  NLI-based semantic sets confidence \citep{Vashurin_2025}.\footnote{Non-contradiction probability replaces $I[y \equiv y_j ]$ with bi-directional average of NLI-estimated $1 - P(\text{contradict})$.} Following previous studies \citep{kuhn2023semanticuncertaintylinguisticinvariances, manakul2023selfcheckgptzeroresourceblackboxhallucination, lin2024generatingconfidenceuncertaintyquantification, Vashurin_2025}, we use \texttt{deberta-large-mnli} for our NLI model \citep{he2021debertadecodingenhancedbertdisentangled}. Figure~\ref{fig:nlg_vs_func} summarizes the results.

The comparison reveals two patterns. For the clustering-based methods (entropy and sets), functional equivalence yields large and consistent improvements across all benchmarks and models. NLI-based semantic entropy and semantic sets confidence perform near chance across all three languages, ranging from 0.47-0.58 on SQL, 0.48-0.56 on Java, and 0.50-0.53 on Python (with one exception at 0.74 on Gemini-2.5-Pro), a consequence of NLI models assigning nearly all responses to a single cluster regardless of functional differences (see Appendix~\ref{sec:sets_dist} for full distributions). The gains from functional equivalence methods are largest on Python and SQL, where the gap exceeds 0.30 AUROC for most models. On Java, functional methods also outperform consistently (0.62-0.70 vs. 0.48-0.56), though the margin is smaller.

For the non-contradiction probability vs.\ functional equivalence rate comparison, the pattern is more nuanced. Non-contradiction often retains a useful continuous signal even when discrete clustering fails, achieving 0.70--0.71 on Python and 0.67--0.79 on SQL. However, functional equivalence rate exceeds it on 12 of 15 benchmark-model combinations, with the largest gains on Python (up to 0.16 AUROC on Gemini-2.5-Flash), while on Java gains are more modest (0.03--0.09). On SQL, the two methods perform comparably, likely because SQL syntax more closely resembles natural language (e.g., \texttt{SELECT height, weight FROM mytable}), allowing the NLI model to partially capture semantic relationships between queries.

\section{Discussion and Conclusion}
\label{sec:discussion}

Our results demonstrate that uncertainty quantification methods can effectively predict functional correctness in LLM-generated code, with functional equivalence methods achieving top AUROC in 11 out of 15 model-benchmark combinations. These methods also exhibit the best calibration across most settings without any explicit calibration procedure, meaning their confidence scores often serve as reliable probability estimates. The comparison with NLI-based counterparts suggests that NLG-specific subcomponents are the primary bottleneck. In particular, NLI-based clustering fails on code because it cannot reliably distinguish functionally different programs, while LLM-based functional equivalence assessment overcomes this limitation directly. This finding has implications beyond our specific methods, suggesting that other NLG-to-code transfer efforts relying on NLI or text embeddings may face similar degradation, and that LLM-based evaluation offers a viable path for domain adaptation of UQ methods more broadly.

\paragraph{Cross-language variation.}
Performance varies substantially across languages. We hypothesize that Python yields the highest AUROC overall due to richer training data and the relative expressiveness of Python solutions, which provide more signal for both token-probability and sampling-based methods. SQL poses the greatest challenge for token-probability methods, which frequently perform near chance, possibly because the constrained syntax of SQL queries leads to high token-level confidence regardless of correctness. Java falls between the two: overall AUROC is lower than Python across all method families, likely reflecting the more rigid boilerplate structure of Java solutions (e.g., class declarations, type annotations), which reduces the diversity of sampled responses and provides less signal for sampling-based methods. 

Sampling-based functional equivalence methods are more robust to these language-level differences, maintaining strong performance across all three settings. Notably, on SQL, GPT-4o-mini achieves 0.84 AUROC with functional entropy despite having the lowest accuracy (0.21), illustrating that sampling-based consistency can rank correct and incorrect generations effectively even for weak models. That said, UQ is most valuable as a complementary signal for a reasonably competent generator; when base accuracy is sufficiently low, the appropriate intervention is model selection rather than uncertainty-based filtering. The cross-language variation in NLI performance is itself informative: NLI-based non-contradiction performs comparably to functional equivalence on SQL, where query syntax is more similar to natural language, but gaps relative to functional equivalence are notably larger on Python and Java, where code structure diverges from natural language patterns.

\paragraph{Equivalence Judge Robustness.} We evaluate the robustness of functional equivalence methods along three axes. First, replacing Gemini-2.5-Flash as the equivalence judge with Gemini-2.5-Flash-Lite produces AUROC within 0.02 across all configurations, while GPT-4o-mini shows moderate degradation (up to 0.07), with functional equivalence methods remaining the top-performing family in most configurations (Appendix~\ref{sec:judge-sensitivity}). Second, a permutation analysis shows that reshuffling the order of sampled responses produces narrow AUROC ranges (typically 2--4 percentage points across 10 runs) and high pairwise Pearson correlations between runs (exceeding 0.94 in 10 of 12 settings), confirming that the greedy clustering procedure is not sensitive to input ordering (Appendix~\ref{app:ordering}). Third, execution-based validation shows cluster purity exceeding 93\% across all settings, indicating low false equivalence. A complementary false non-equivalence analysis finds moderate response-level misclassification rates for Python (10.3--17.9\%) and higher upper-bound rates for SQL (42.4--45.1\%), though the resulting entropy inflation is conservative (overestimating rather than underestimating uncertainty) and does not degrade AUROC (Appendix~\ref{sec:validation}).

\paragraph{Practitioner recommendations.}
For practitioners, we recommend functional equivalence methods when the sampling and equivalence assessment costs are acceptable, as these methods generally perform best in classification and calibration. When sampling is not feasible, whether due to latency constraints or API costs, minimum token negentropy is the strongest single-generation alternative across benchmarks, achieving highest within-family AUROC in 12 of 15 model-benchmark combinations. Even at $m=10$ generations, equivalence scoring costs are modest relative to the sampled generation cost that all sampling-based methods share (Appendix~\ref{sec:cost}). Moreover, our ablation in Appendix~\ref{sec:sampled_ablation} shows that $m=5$ captures most of the AUROC gain (within 0.01--0.02 of $m=10$ across most settings), reducing functional equivalence rate to 5 judge calls per prompt and clustering-based methods to at most 15.

\paragraph{Execution-based confidence.}
Dynamic execution remains the preferred validation signal when high-quality tests and a reliable execution environment are available. An alternative confidence pipeline is to generate test cases, execute each sampled program, and estimate uncertainty from pass rates or behavioral agreement. This can provide a more direct measure of functional correctness than static equivalence judgment, but it also introduces additional cost and failure modes, as test generation requires extra model calls, execution adds latency and sandboxing overhead, and the resulting confidence depends on test quality and coverage. Generated tests may have limited coverage or correlate with the model's own solution patterns, so passing generated tests does not guarantee correctness. Functional entropy occupies a different point in this trade-off space. It is execution-free, applies when dependencies, sandboxes, private databases, or complete test harnesses are unavailable, and can be used uniformly across Python, Java, and SQL. Its limitation is that it relies on an approximate equivalence judge, which can miscluster programs. We view execution-based confidence as preferable when reliable execution is available, and functional entropy as a complementary execution-free alternative when sampling is feasible but execution is impractical.

\paragraph{Future work.}
All methods evaluated in this paper are available 
through \texttt{uqlm}, our
open-source library for LLM uncertainty quantification. Future work includes extending evaluation to additional programming languages and models, as well as investigating the sensitivity of sampling-based methods to temperature and top-p hyperparameters. Several directions could improve equivalence assessment: language-specific prompt-optimization, code-structure-aware approaches using abstract syntax trees or control flow graphs, symbolic execution for classes of programs where it is tractable, and domain-specialized judge models. Lightweight execution proxies such as property-based testing or LLM-generated mini-tests represent a promising middle ground between fully execution-free and full test-suite-based approaches. In particular, the judge could identify candidate equivalence clusters while selectively generated tests verify uncertain judgments. Finally, recent work on Semantic Entropy Probes \citep{kossen2024semanticentropyprobesrobust}, which train lightweight models to predict semantic entropy from model internals, could in principle be extended to functional entropy, potentially eliminating the need for both sampling and pairwise judge calls at inference time.


\section*{Limitations}

\paragraph{Coverage of benchmarks and programming languages.} We evaluate on three benchmarks covering Python, Java, and SQL, which represent a subset of programming languages and task types. Our findings may not generalize to other languages with different syntactic properties (e.g., statically typed languages like Rust or Haskell), to longer code generation tasks involving multi-file projects, or to tasks requiring interaction with external APIs or databases. LiveSQLBench contains only 270 problems, which limits statistical power for SQL-specific conclusions. In particular, cross-method comparisons on SQL should be interpreted with more caution than those on Python (1,055 problems) or Java (386 problems), and claims about SQL-specific patterns, such as the relative competitiveness of NLI-based methods on SQL syntax, would benefit from validation on larger benchmarks as they become available.

\paragraph{Model coverage.} All five models evaluated are closed-source API models from two providers (OpenAI and Google). Other models may differ in calibration properties and generation characteristics. In particular, our evaluation does not include open-weight models (e.g., Llama, Mistral, Qwen), which are increasingly common in cost-sensitive deployment settings. Our results should not be assumed to transfer directly to these or other LLMs without further evaluation.

\paragraph{Single-turn scope.} Our evaluation targets single-turn code generation on established benchmarks. Extending to multi-turn agentic settings, where generated code is iteratively refined or executed as part of a broader tool-use pipeline, is a natural next step but may introduce additional challenges such as error propagation across turns.

\paragraph{Computational cost.} Sampling-based methods require generating multiple responses per prompt and, in the case of functional equivalence methods, additional LLM calls for pairwise equivalence assessment. This increases both latency and cost relative to single-generation token-probability methods. We report dollar costs in Appendix~\ref{sec:cost} but do not measure wall-clock latency, as it is dominated by API-specific factors (e.g., rate limits, queue depth) that vary across providers and deployment settings.

\paragraph{Correctness definition.} We define functional correctness using test-based evaluation (i.e., a solution is correct if it passes all provided test cases). This is an approximation, as test suites may not cover all edge cases, and a solution that passes all tests may still contain latent bugs. This limitation applies to UQ methods for both Python and Java.

\paragraph{Sampling hyperparameters.} We fix temperature at $T=1.0$ across all models and leave sensitivity to sampling hyperparameters to future work.

\paragraph{Equivalence judge limitations.} Functional equivalence is undecidable in general (Rice's Theorem). Our LLM-based judge is an approximate classifier that can exhibit both false equivalence and false non-equivalence (Appendix~\ref{sec:validation}). It is not intended to replace execution-based validation when reliable tests and a runtime environment are available.



\section*{Ethical Considerations}
While our research enhances the reliability of AI-generated code, several ethical considerations remain. First, uncertainty quantification (UQ) is not a proxy for security; code with low functional entropy may still contain latent bugs or vulnerabilities, making continued human oversight and security auditing essential. Second, the increased computational demand of sampling-based functional equivalence methods poses environmental concerns and may limit accessibility for researchers in resource-constrained settings compared to more efficient token-probability alternatives. Finally, our findings are based on a specific set of models and languages, and practitioners should exercise caution when generalizing these results to other programming domains or architectures without further validation.

\section*{Disclosure}
\paragraph{Conflict of interest.} MSC, ZA, and HKR are employed by, and receive stock and equity from, CVS Health\textsuperscript{\textregistered} Corporation. DB was employed by, and received stock and equity from, CVS Health\textsuperscript{\textregistered} Corporation at the time this work was conducted.

\paragraph{Disclaimer.} Prompts are included solely for reproducibility and do not imply endorsement or affiliation. Gemini is a trademark of Google and GPT is a trademark of OpenAI. This is an independent publication and has not been authorized, endorsed, or sponsored by Google or OpenAI.

\paragraph{LLM usage.} The authors used LLMs to assist with editing the manuscript.

\bibliography{refs}

\appendix

\section{Calibration}
\label{sec:calibration}
We report two calibration metrics for all methods evaluated in the main text: Expected Calibration Error (ECE) and Brier Score. ECE measures the average gap between predicted confidence and observed accuracy across binned confidence intervals, while Brier Score measures the mean squared error between confidence scores and binary correctness labels. For both metrics, lower values indicate better calibration. Tables~\ref{tab:ECE_LiveCod}, \ref{tab:ECE_MultiPL}, and \ref{tab:ECE_LiveSQL} report ECE results, while Tables~\ref{tab:Brier Score_LiveCod}, \ref{tab:Brier Score_MultiPL}, and \ref{tab:Brier Score_LiveSQL} report Brier Score across all three benchmarks.

The calibration results are broadly consistent with the AUROC findings reported in the main text. Functional equivalence methods (functional entropy, functional sets confidence) achieve the best calibration across most benchmark-model combinations, with ECE values frequently below 0.20 and competitive Brier scores. This is particularly notable on Python problems, where these methods produce well-calibrated confidence estimates without any explicit calibration procedure.

Token-probability methods show a wide calibration range. Minimum token negentropy and minimum probability tend to be better calibrated than sequence probability, mean token negentropy, and probability margin, which often exhibit high ECE, particularly on SQL where values exceed 0.70 in several cases. This pattern is consistent with the AUROC finding that aggregated token-probability methods are often confidently wrong on SQL.

Normalized cosine similarity, despite reasonable AUROC performance, shows poor calibration on SQL and Java, with ECE and Brier scores comparable to the weakest token-probability methods. This suggests that embedding-based similarity scores do not naturally align with correctness probabilities, even when they provide useful ranking signal.

Among reflexive methods, P(True) shows variable calibration across benchmarks, while verbalized confidence is generally poorly calibrated, consistent with prior observations that LLMs tend to express overconfident or inconsistent self-assessments.



\begin{table}[htp]
\tiny
\centering
\begin{tabular}{lrrrrr}
\toprule
Scorer & Gem-Fl & Gem-Pro & Gem-Fl-Lt & GPT4o-m & GPT4o \\
\midrule
\multicolumn{6}{l}{\textit{Token-Probability Methods}} \\
Seq. Prob. & \cellcolor{blue!35!white}0.051 & \cellcolor{blue!45!white}\underline{0.015} & \cellcolor{red!5!white}0.253 & \cellcolor{red!25!white}0.352 & \cellcolor{red!5!white}0.220 \\
Min. Prob. & \cellcolor{red!45!white}0.726 & \cellcolor{red!45!white}0.824 & \cellcolor{red!45!white}0.502 & \cellcolor{red!35!white}0.390 & \cellcolor{red!45!white}0.501 \\
Prob. Margin & \cellcolor{blue!35!white}0.059 & \cellcolor{blue!45!white}0.022 & \cellcolor{red!15!white}0.262 & \cellcolor{red!35!white}0.414 & \cellcolor{red!25!white}0.325 \\
Avg. Tok. Neg. & \cellcolor{blue!15!white}0.091 & \cellcolor{blue!35!white}0.050 & \cellcolor{red!15!white}0.292 & \cellcolor{red!35!white}0.444 & \cellcolor{red!25!white}0.366 \\
Min. Tok. Neg. & \cellcolor{red!45!white}0.535 & \cellcolor{red!45!white}0.538 & \cellcolor{red!15!white}0.271 & \cellcolor{blue!25!white}0.082 & \cellcolor{red!5!white}0.251 \\
\midrule
\multicolumn{6}{l}{\textit{Functional Equivalence Methods}} \\
Func. Equiv. Rate & \cellcolor{blue!5!white}0.110 & \cellcolor{red!5!white}0.200 & \cellcolor{blue!5!white}0.113 & \cellcolor{blue!15!white}0.100 & \cellcolor{blue!5!white}0.117 \\
Func. Neg. & \cellcolor{blue!25!white}0.073 & \cellcolor{blue!5!white}0.152 & \cellcolor{blue!45!white}\underline{0.038} & \cellcolor{blue!45!white}\underline{0.041} & \cellcolor{blue!35!white}\underline{0.046} \\
Func. Sets & \cellcolor{blue!45!white}\underline{0.024} & \cellcolor{blue!25!white}0.080 & \cellcolor{blue!25!white}0.071 & \cellcolor{blue!5!white}0.110 & \cellcolor{blue!15!white}0.090 \\
\midrule
\multicolumn{6}{l}{\textit{Similarity-Based Methods}} \\
Cos. Sim. & \cellcolor{blue!35!white}0.044 & \cellcolor{blue!45!white}0.039 & \cellcolor{red!15!white}0.273 & \cellcolor{red!35!white}0.430 & \cellcolor{red!25!white}0.362 \\
CodeBLEU & \cellcolor{red!35!white}0.428 & \cellcolor{red!35!white}0.468 & \cellcolor{blue!15!white}0.085 & \cellcolor{blue!25!white}0.082 & \cellcolor{blue!15!white}0.100 \\
\midrule
\multicolumn{6}{l}{\textit{Reflexive Methods}} \\
Verb. Conf. & \cellcolor{blue!25!white}0.072 & \cellcolor{blue!35!white}0.060 & \cellcolor{red!5!white}0.260 & \cellcolor{red!15!white}0.281 & \cellcolor{red!25!white}0.303 \\
P(True) & \cellcolor{blue!5!white}0.111 & \cellcolor{blue!15!white}0.087 & \cellcolor{red!5!white}0.210 & \cellcolor{red!25!white}0.349 & \cellcolor{red!15!white}0.300 \\
\bottomrule
\end{tabular}
\caption{ECE for predicting code correctness on LiveCodeBench (Python). Lower is better. Colors reflect rankings: blue (best) to red (worst). The best-calibrated scorer based on ECE is \underline{underlined}.}
\label{tab:ECE_LiveCod}
\end{table}

\begin{table}[htp]
\tiny
\centering
\begin{tabular}{lrrrrr}
\toprule
Scorer & Gem-Fl & Gem-Pro & Gem-Fl-Lt & GPT4o-m & GPT4o \\
\midrule
\multicolumn{6}{l}{\textit{Token-Probability Methods}} \\
Seq. Prob. & \cellcolor{red!25!white}0.431 & \cellcolor{red!35!white}0.435 & \cellcolor{red!45!white}0.465 & \cellcolor{red!5!white}0.374 & \cellcolor{red!5!white}0.359 \\
Min. Prob. & \cellcolor{blue!15!white}0.244 & \cellcolor{blue!5!white}0.249 & \cellcolor{blue!25!white}0.211 & \cellcolor{blue!5!white}0.289 & \cellcolor{blue!5!white}0.300 \\
Prob. Margin & \cellcolor{red!35!white}0.433 & \cellcolor{red!35!white}0.432 & \cellcolor{red!45!white}0.468 & \cellcolor{red!15!white}0.386 & \cellcolor{red!15!white}0.375 \\
Avg. Tok. Neg. & \cellcolor{red!45!white}0.463 & \cellcolor{red!35!white}0.451 & \cellcolor{red!45!white}0.493 & \cellcolor{red!15!white}0.413 & \cellcolor{red!15!white}0.400 \\
Min. Tok. Neg. & \cellcolor{blue!35!white}0.123 & \cellcolor{blue!35!white}0.164 & \cellcolor{blue!45!white}\underline{0.075} & \cellcolor{blue!45!white}\underline{0.105} & \cellcolor{blue!45!white}0.099 \\
\midrule
\multicolumn{6}{l}{\textit{Functional Equivalence Methods}} \\
Func. Equiv. Rate & \cellcolor{blue!15!white}0.241 & \cellcolor{blue!25!white}0.212 & \cellcolor{blue!15!white}0.222 & \cellcolor{blue!15!white}0.221 & \cellcolor{blue!25!white}0.198 \\
Func. Neg. & \cellcolor{blue!35!white}0.189 & \cellcolor{blue!25!white}0.206 & \cellcolor{blue!15!white}0.223 & \cellcolor{blue!35!white}0.197 & \cellcolor{blue!35!white}0.155 \\
Func. Sets & \cellcolor{blue!5!white}0.294 & \cellcolor{blue!5!white}0.297 & \cellcolor{red!5!white}0.308 & \cellcolor{blue!5!white}0.262 & \cellcolor{blue!15!white}0.229 \\
\midrule
\multicolumn{6}{l}{\textit{Similarity-Based Methods}} \\
Cos. Sim. & \cellcolor{red!35!white}0.448 & \cellcolor{red!25!white}0.420 & \cellcolor{red!45!white}0.486 & \cellcolor{red!25!white}0.414 & \cellcolor{red!15!white}0.394 \\
CodeBLEU & \cellcolor{blue!45!white}\underline{0.089} & \cellcolor{blue!45!white}\underline{0.086} & \cellcolor{blue!25!white}0.209 & \cellcolor{blue!35!white}0.149 & \cellcolor{blue!45!white}\underline{0.084} \\
\midrule
\multicolumn{6}{l}{\textit{Reflexive Methods}} \\
Verb. Conf. & \cellcolor{red!15!white}0.375 & \cellcolor{red!25!white}0.425 & \cellcolor{red!35!white}0.447 & \cellcolor{blue!25!white}0.220 & \cellcolor{red!5!white}0.342 \\
P(True) & \cellcolor{red!45!white}0.466 & \cellcolor{red!25!white}0.424 & \cellcolor{red!25!white}0.416 & \cellcolor{red!5!white}0.350 & \cellcolor{red!5!white}0.324 \\
\bottomrule
\end{tabular}
\caption{ECE for predicting code correctness on MultiPL-E (Java). Lower is better. Colors reflect rankings: blue (best) to red (worst). The best-calibrated scorer based on ECE is \underline{underlined}.}
\label{tab:ECE_MultiPL}
\end{table}

\begin{table}[htp]
\tiny
\centering
\begin{tabular}{lrrrrr}
\toprule
Scorer & Gem-Fl & Gem-Pro & Gem-Fl-Lt & GPT4o-m & GPT4o \\
\midrule
\multicolumn{6}{l}{\textit{Token-Probability Methods}} \\
Seq. Prob. & \cellcolor{red!25!white}0.523 & \cellcolor{red!15!white}0.521 & \cellcolor{red!35!white}0.714 & \cellcolor{red!35!white}0.675 & \cellcolor{red!5!white}0.417 \\
Min. Prob. & \cellcolor{blue!5!white}0.260 & \cellcolor{blue!15!white}0.251 & \cellcolor{blue!25!white}0.150 & \cellcolor{blue!35!white}0.144 & \cellcolor{blue!5!white}0.260 \\
Prob. Margin & \cellcolor{red!25!white}0.523 & \cellcolor{red!15!white}0.520 & \cellcolor{red!35!white}0.717 & \cellcolor{red!45!white}0.724 & \cellcolor{red!35!white}0.592 \\
Avg. Tok. Neg. & \cellcolor{red!25!white}0.533 & \cellcolor{red!25!white}0.533 & \cellcolor{red!45!white}0.738 & \cellcolor{red!45!white}0.752 & \cellcolor{red!35!white}0.633 \\
Min. Tok. Neg. & \cellcolor{blue!35!white}\underline{0.126} & \cellcolor{blue!25!white}0.164 & \cellcolor{blue!15!white}0.180 & \cellcolor{blue!15!white}0.257 & \cellcolor{blue!45!white}\underline{0.040} \\
\midrule
\multicolumn{6}{l}{\textit{Functional Equivalence Methods}} \\
Func. Equiv. Rate & \cellcolor{blue!5!white}0.262 & \cellcolor{blue!15!white}0.208 & \cellcolor{blue!25!white}0.154 & \cellcolor{blue!35!white}0.124 & \cellcolor{blue!25!white}0.177 \\
Func. Neg. & \cellcolor{blue!15!white}0.192 & \cellcolor{blue!45!white}0.110 & \cellcolor{blue!35!white}\underline{0.130} & \cellcolor{blue!45!white}\underline{0.067} & \cellcolor{blue!35!white}0.119 \\
Func. Sets & \cellcolor{blue!25!white}0.157 & \cellcolor{blue!45!white}\underline{0.102} & \cellcolor{blue!25!white}0.146 & \cellcolor{blue!45!white}0.083 & \cellcolor{blue!45!white}0.089 \\
\midrule
\multicolumn{6}{l}{\textit{Similarity-Based Methods}} \\
Cos. Sim. & \cellcolor{red!15!white}0.516 & \cellcolor{red!15!white}0.513 & \cellcolor{red!45!white}0.727 & \cellcolor{red!45!white}0.756 & \cellcolor{red!35!white}0.648 \\
\midrule
\multicolumn{6}{l}{\textit{Reflexive Methods}} \\
Verb. Conf. & \cellcolor{blue!5!white}0.412 & \cellcolor{red!5!white}0.464 & \cellcolor{red!5!white}0.441 & \cellcolor{red!15!white}0.509 & \cellcolor{red!25!white}0.573 \\
P(True) & \cellcolor{red!5!white}0.485 & \cellcolor{red!5!white}0.423 & \cellcolor{blue!5!white}0.400 & \cellcolor{red!15!white}0.496 & \cellcolor{blue!5!white}0.323 \\
\bottomrule
\end{tabular}
\caption{ECE for predicting code correctness on LiveSQLBench (SQLite). Lower is better. Colors reflect rankings: blue (best) to red (worst). The best-calibrated scorer based on ECE is \underline{underlined}.}
\label{tab:ECE_LiveSQL}
\end{table}



\begin{table}[htp]
\tiny
\centering
\begin{tabular}{lrrrrr}
\toprule
Scorer & Gem-Fl & Gem-Pro & Gem-Fl-Lt & GPT4o-m & GPT4o \\
\midrule
\multicolumn{6}{l}{\textit{Token-Probability Methods}} \\
Seq. Prob. & \cellcolor{blue!25!white}0.109 & \cellcolor{blue!45!white}0.078 & \cellcolor{red!5!white}0.263 & \cellcolor{red!25!white}0.348 & \cellcolor{red!5!white}0.258 \\
Min. Prob. & \cellcolor{red!45!white}0.678 & \cellcolor{red!45!white}0.767 & \cellcolor{red!45!white}0.449 & \cellcolor{red!35!white}0.373 & \cellcolor{red!45!white}0.485 \\
Prob. Margin & \cellcolor{blue!25!white}0.112 & \cellcolor{blue!45!white}0.079 & \cellcolor{red!15!white}0.272 & \cellcolor{red!35!white}0.406 & \cellcolor{red!25!white}0.332 \\
Avg. Tok. Neg. & \cellcolor{blue!25!white}0.118 & \cellcolor{blue!45!white}0.081 & \cellcolor{red!15!white}0.294 & \cellcolor{red!45!white}0.436 & \cellcolor{red!35!white}0.365 \\
Min. Tok. Neg. & \cellcolor{red!35!white}0.416 & \cellcolor{red!35!white}0.371 & \cellcolor{red!5!white}0.244 & \cellcolor{blue!5!white}0.198 & \cellcolor{red!5!white}0.263 \\
\midrule
\multicolumn{6}{l}{\textit{Functional Equivalence Methods}} \\
Func. Equiv. Rate & \cellcolor{blue!25!white}0.108 & \cellcolor{blue!15!white}0.151 & \cellcolor{blue!5!white}0.161 & \cellcolor{blue!15!white}0.161 & \cellcolor{blue!5!white}0.165 \\
Func. Neg. & \cellcolor{blue!35!white}0.092 & \cellcolor{blue!35!white}0.104 & \cellcolor{blue!25!white}\underline{0.145} & \cellcolor{blue!15!white}\underline{0.152} & \cellcolor{blue!15!white}\underline{0.153} \\
Func. Sets & \cellcolor{blue!35!white}\underline{0.083} & \cellcolor{blue!45!white}0.073 & \cellcolor{blue!15!white}0.150 & \cellcolor{blue!5!white}0.166 & \cellcolor{blue!15!white}0.161 \\
\midrule
\multicolumn{6}{l}{\textit{Similarity-Based Methods}} \\
Cos. Sim. & \cellcolor{blue!35!white}0.108 & \cellcolor{blue!45!white}0.073 & \cellcolor{red!15!white}0.281 & \cellcolor{red!45!white}0.420 & \cellcolor{red!35!white}0.363 \\
CodeBLEU & \cellcolor{red!15!white}0.289 & \cellcolor{red!25!white}0.294 & \cellcolor{blue!5!white}0.179 & \cellcolor{blue!5!white}0.203 & \cellcolor{red!5!white}0.219 \\
\midrule
\multicolumn{6}{l}{\textit{Reflexive Methods}} \\
Verb. Conf. & \cellcolor{blue!35!white}0.090 & \cellcolor{blue!45!white}\underline{0.067} & \cellcolor{red!15!white}0.279 & \cellcolor{red!25!white}0.295 & \cellcolor{red!25!white}0.302 \\
P(True) & \cellcolor{blue!25!white}0.111 & \cellcolor{blue!35!white}0.086 & \cellcolor{red!5!white}0.212 & \cellcolor{red!25!white}0.348 & \cellcolor{red!15!white}0.290 \\
\bottomrule
\end{tabular}
\caption{Brier Score for predicting code correctness on LiveCodeBench (Python). Lower is better. Colors reflect rankings: blue (best) to red (worst). The best-calibrated scorer based on Brier Score is \underline{underlined}.}
\label{tab:Brier Score_LiveCod}
\end{table}

\begin{table}[htp]
\tiny
\centering
\begin{tabular}{lrrrrr}
\toprule
Scorer & Gem-Fl & Gem-Pro & Gem-Fl-Lt & GPT4o-m & GPT4o \\
\midrule
\multicolumn{6}{l}{\textit{Token-Probability Methods}} \\
Seq. Prob. & \cellcolor{red!35!white}0.434 & \cellcolor{red!35!white}0.435 & \cellcolor{red!45!white}0.460 & \cellcolor{red!15!white}0.383 & \cellcolor{red!5!white}0.364 \\
Min. Prob. & \cellcolor{blue!5!white}0.306 & \cellcolor{blue!5!white}0.303 & \cellcolor{blue!35!white}0.263 & \cellcolor{red!5!white}0.340 & \cellcolor{blue!5!white}0.325 \\
Prob. Margin & \cellcolor{red!35!white}0.438 & \cellcolor{red!25!white}0.432 & \cellcolor{red!45!white}0.465 & \cellcolor{red!15!white}0.394 & \cellcolor{red!5!white}0.378 \\
Avg. Tok. Neg. & \cellcolor{red!45!white}0.463 & \cellcolor{red!35!white}0.450 & \cellcolor{red!45!white}0.490 & \cellcolor{red!25!white}0.415 & \cellcolor{red!15!white}0.400 \\
Min. Tok. Neg. & \cellcolor{blue!35!white}0.250 & \cellcolor{blue!25!white}0.263 & \cellcolor{blue!45!white}\underline{0.232} & \cellcolor{blue!35!white}\underline{0.245} & \cellcolor{blue!45!white}0.237 \\
\midrule
\multicolumn{6}{l}{\textit{Functional Equivalence Methods}} \\
Func. Equiv. Rate & \cellcolor{blue!15!white}0.303 & \cellcolor{blue!15!white}0.280 & \cellcolor{blue!25!white}0.274 & \cellcolor{blue!25!white}0.273 & \cellcolor{blue!35!white}0.259 \\
Func. Neg. & \cellcolor{blue!15!white}0.284 & \cellcolor{blue!15!white}0.275 & \cellcolor{blue!25!white}0.272 & \cellcolor{blue!35!white}0.254 & \cellcolor{blue!45!white}0.243 \\
Func. Sets & \cellcolor{blue!5!white}0.326 & \cellcolor{blue!5!white}0.314 & \cellcolor{blue!5!white}0.314 & \cellcolor{blue!15!white}0.283 & \cellcolor{blue!25!white}0.266 \\
\midrule
\multicolumn{6}{l}{\textit{Similarity-Based Methods}} \\
Cos. Sim. & \cellcolor{red!35!white}0.445 & \cellcolor{red!25!white}0.419 & \cellcolor{red!45!white}0.484 & \cellcolor{red!15!white}0.413 & \cellcolor{red!15!white}0.396 \\
CodeBLEU & \cellcolor{blue!45!white}\underline{0.231} & \cellcolor{blue!45!white}\underline{0.236} & \cellcolor{blue!25!white}0.271 & \cellcolor{blue!35!white}0.252 & \cellcolor{blue!45!white}\underline{0.233} \\
\midrule
\multicolumn{6}{l}{\textit{Reflexive Methods}} \\
Verb. Conf. & \cellcolor{red!15!white}0.379 & \cellcolor{red!25!white}0.425 & \cellcolor{red!35!white}0.438 & \cellcolor{blue!15!white}0.282 & \cellcolor{red!5!white}0.354 \\
P(True) & \cellcolor{red!45!white}0.466 & \cellcolor{red!25!white}0.423 & \cellcolor{red!25!white}0.419 & \cellcolor{red!5!white}0.360 & \cellcolor{red!5!white}0.331 \\
\bottomrule
\end{tabular}
\caption{Brier Score for predicting code correctness on MultiPL-E (Java). Lower is better. Colors reflect rankings: blue (best) to red (worst). The best-calibrated scorer based on Brier Score is \underline{underlined}.}
\label{tab:Brier Score_MultiPL}
\end{table}

\begin{table}[htp]
\tiny
\centering
\begin{tabular}{lrrrrr}
\toprule
Scorer & Gem-Fl & Gem-Pro & Gem-Fl-Lt & GPT4o-m & GPT4o \\
\midrule
\multicolumn{6}{l}{\textit{Token-Probability Methods}} \\
Seq. Prob. & \cellcolor{red!25!white}0.520 & \cellcolor{red!15!white}0.519 & \cellcolor{red!35!white}0.687 & \cellcolor{red!35!white}0.627 & \cellcolor{red!5!white}0.398 \\
Min. Prob. & \cellcolor{blue!5!white}0.302 & \cellcolor{blue!5!white}0.308 & \cellcolor{blue!35!white}0.199 & \cellcolor{blue!35!white}0.175 & \cellcolor{blue!15!white}0.266 \\
Prob. Margin & \cellcolor{red!25!white}0.521 & \cellcolor{red!15!white}0.518 & \cellcolor{red!45!white}0.692 & \cellcolor{red!35!white}0.686 & \cellcolor{red!35!white}0.553 \\
Avg. Tok. Neg. & \cellcolor{red!25!white}0.532 & \cellcolor{red!25!white}0.532 & \cellcolor{red!45!white}0.724 & \cellcolor{red!45!white}0.729 & \cellcolor{red!35!white}0.604 \\
Min. Tok. Neg. & \cellcolor{blue!25!white}\underline{0.248} & \cellcolor{blue!5!white}0.266 & \cellcolor{blue!25!white}0.200 & \cellcolor{blue!25!white}0.225 & \cellcolor{blue!35!white}0.184 \\
\midrule
\multicolumn{6}{l}{\textit{Functional Equivalence Methods}} \\
Func. Equiv. Rate & \cellcolor{blue!5!white}0.302 & \cellcolor{blue!15!white}0.263 & \cellcolor{blue!45!white}0.169 & \cellcolor{blue!45!white}0.147 & \cellcolor{blue!25!white}0.200 \\
Func. Neg. & \cellcolor{blue!15!white}0.261 & \cellcolor{blue!25!white}0.215 & \cellcolor{blue!45!white}\underline{0.168} & \cellcolor{blue!45!white}\underline{0.129} & \cellcolor{blue!35!white}\underline{0.183} \\
Func. Sets & \cellcolor{blue!15!white}\underline{0.248} & \cellcolor{blue!25!white}\underline{0.207} & \cellcolor{blue!35!white}0.176 & \cellcolor{blue!45!white}0.135 & \cellcolor{blue!45!white}0.174 \\
\midrule
\multicolumn{6}{l}{\textit{Similarity-Based Methods}} \\
Cos. Sim. & \cellcolor{red!15!white}0.510 & \cellcolor{red!15!white}0.503 & \cellcolor{red!45!white}0.705 & \cellcolor{red!45!white}0.734 & \cellcolor{red!35!white}0.620 \\
\midrule
\multicolumn{6}{l}{\textit{Reflexive Methods}} \\
Verb. Conf. & \cellcolor{red!5!white}0.430 & \cellcolor{red!5!white}0.460 & \cellcolor{blue!5!white}0.368 & \cellcolor{red!5!white}0.431 & \cellcolor{red!25!white}0.548 \\
P(True) & \cellcolor{red!15!white}0.485 & \cellcolor{red!5!white}0.422 & \cellcolor{blue!5!white}0.381 & \cellcolor{red!15!white}0.477 & \cellcolor{blue!15!white}0.264 \\
\bottomrule
\end{tabular}
\caption{Brier Score for predicting code correctness on LiveSQLBench (SQLite). Lower is better. Colors reflect rankings: blue (best) to red (worst). The best-calibrated scorer based on Brier Score is \underline{underlined}.}
\label{tab:Brier Score_LiveSQL}
\end{table}


\section{Hybrid Methods}
\label{sec:hybrid}

Hybrid methods combine token-probability signals with sampling-based consistency measures, leveraging both the model's internal confidence and agreement across multiple generations. We evaluate three hybrid methods described below. While these methods are reasonably competitive, none achieve the highest AUROC on any benchmark-model combination in our experiments, and we therefore present these results in the appendix rather than the main text.

\subsection{Methods}

\paragraph{Consistency and Confidence Approach (CoCoA).} CoCoA combines the length-normalized sequence probability of the original response with embedding-based consistency \citep{vashurin2025uncertaintyquantificationllmsminimum}. Let $y_0$ denote the original response and $\{\tilde{y}_1, \ldots, \tilde{y}_m\}$ denote the sampled responses. CoCoA is defined as:
\begin{equation*}
    \text{CoCoA}(y; \tilde{\mathbf{y}}, x) = \text{LNSP}(y) \cdot \text{NCS}(y; \tilde{\mathbf{y}})
\end{equation*}
We use the code embedding variant described in Section~\ref{sec:consistency}.

\paragraph{Monte Carlo Sequence Probability (MCSP).} This method computes the average length-normalized sequence probability across $m$ sampled responses $\{y_1, \ldots, y_m\}$ \citep{kuhn2023semanticuncertaintylinguisticinvariances}:
\begin{equation*}
    \text{MCSP}(y; \tilde{\mathbf{y}}, x) = \frac{1}{m+1} \sum_{i=0}^{m} \text{LNSP}(y_i)
\end{equation*}
where $y_0 = y$ denotes the original response.

\paragraph{White-Box Normalized Functional Negentropy (WB-NFN).} In the hybrid setting, we weight cluster probabilities by the length-normalized sequence probabilities of responses within each cluster, rather than treating all responses equally:
\begin{equation*}
    P(C) = \frac{\sum_{y_i \in C} \text{LNSP}(y_i)}{\sum_{i=0}^{m} \text{LNSP}(y_i)}
\end{equation*}
This white-box version replaces the frequency-based $P(C)$ used in the black-box NFN (Section~\ref{sec:consistency}) with a token-probability-weighted version. The remaining calculations, including LLM-based functional equivalence for clustering, are identical.

\subsection{Results}

Tables~\ref{tab:hybrid_python}, \ref{tab:hybrid_java}, and \ref{tab:hybrid_sql} present AUROC results for hybrid methods across all three benchmarks.

Across all three benchmarks, no hybrid method achieves the top AUROC for any model. WB-NFN is the strongest of the three, reaching 0.850 on Python (Gemini-2.5-Flash) and 0.817 on SQL (GPT-4o-mini), but these scores are consistently matched or exceeded by the black-box functional entropy methods reported in the main text. This suggests that incorporating token-probability weighting into the cluster distribution does not provide additional discriminative signal beyond what frequency-based functional equivalence clustering already captures. MCSP shows high variance across models, performing well on Python for some models (0.821 on GPT-4o) but poorly on Java (0.538 on Gemini-2.5-Flash). CoCoA performs moderately across all benchmarks without distinguishing itself on any, indicating that the multiplicative combination of sequence probability and embedding similarity does not yield complementary signal for code generation.

\begin{table}[t]
\centering
\tiny
\begin{tabular}{lccccc}
\toprule
Scorer & Gem-Fl & Gem-Pro & Gem-Fl-Lt & GPT4o-m & GPT4o \\
\midrule
MCSP & 0.718 & 0.594 & 0.809 & 0.808 & 0.821 \\
WB-NFN & 0.850 & 0.802 & 0.787 & 0.789 & 0.825 \\
CoCoA & 0.783 & 0.793 & 0.790 & 0.772 & 0.746 \\
\bottomrule
\end{tabular}
\caption{Hybrid method AUROC on LiveCodeBench (Python).}
\label{tab:hybrid_python}
\end{table}

\begin{table}[t]
\centering
\tiny
\begin{tabular}{lccccc}
\toprule
Scorer & Gem-Fl & Gem-Pro & Gem-Fl-Lt & GPT4o-m & GPT4o \\
\midrule
MCSP & 0.538 & 0.588 & 0.585 & 0.557 & 0.645 \\
WB-NFN & 0.626 & 0.661 & 0.697 & 0.699 & 0.686 \\
CoCoA & 0.605 & 0.621 & 0.628 & 0.584 & 0.643 \\
\bottomrule
\end{tabular}
\caption{Hybrid method AUROC on MultiPL-E (Java).}
\label{tab:hybrid_java}
\end{table}

\begin{table}[t]
\tiny
\centering
\begin{tabular}{lccccc}
\toprule
Scorer & Gem-Fl & Gem-Pro & Gem-Fl-Lt & GPT4o-m & GPT4o \\
\midrule
MCSP & 0.559 & 0.632 & 0.654 & 0.719 & 0.748 \\
WB-NFN & 0.672 & 0.748 & 0.732 & 0.817 & 0.624 \\
CoCoA & 0.627 & 0.676 & 0.683 & 0.706 & 0.674 \\
\bottomrule
\end{tabular}
\caption{Hybrid method AUROC on LiveSQLBench (SQL).}
\label{tab:hybrid_sql}
\end{table}

\section{Semantic and Functional Sets Distributions}
\label{sec:sets_dist}

\begin{figure*}[tb]
    \centering
    \includegraphics[width=\textwidth]{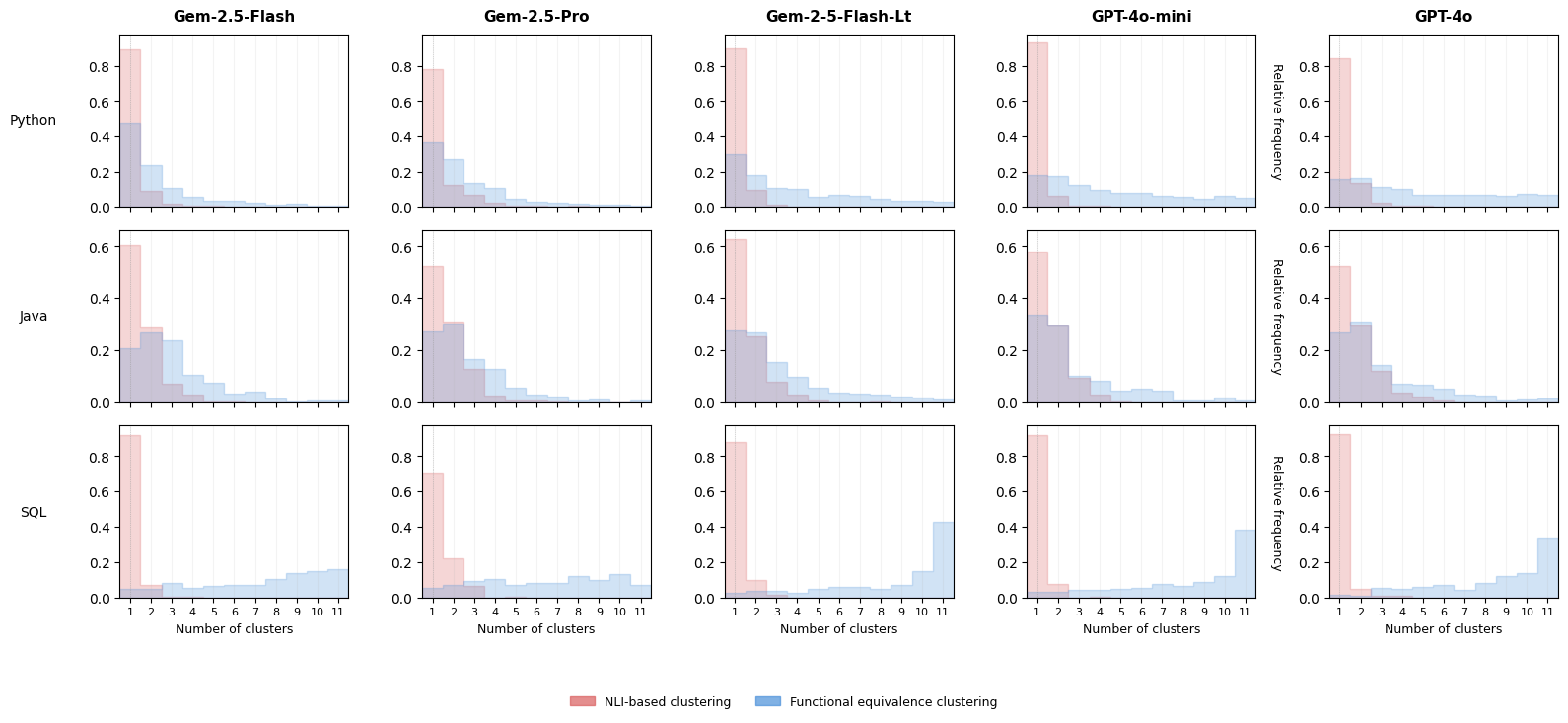}
    \caption{Distribution of cluster counts for NLI-based clustering (red) and functional equivalence clustering (blue) across all models and languages. NLI-based clustering collapses nearly all responses into a single cluster, while functional equivalence clustering produces meaningful variation. Each subplot shows the relative frequency of cluster counts across all prompts for a given model-language combination, with $m=10$ sampled responses per prompt (yielding a maximum of 11 clusters).}
    \label{fig:cluster_distributions}
\end{figure*}

Figure~\ref{fig:cluster_distributions} compares the distribution of cluster counts produced by NLI-based clustering and functional equivalence clustering across all models and languages. Overall, our results indicate that the number of clusters directly determines the behavior of functional entropy and functional sets confidence. In particular, when all responses collapse into a single cluster, entropy is zero and sets confidence is maximal, regardless of whether the underlying code is correct.

NLI-based clustering assigns the vast majority of responses to a single cluster across all settings, with over 80\% of prompts producing exactly one cluster on Python and SQL. This collapse explains why NLI-based semantic entropy and semantic sets confidence perform near chance in the main experiments, i.e., the resulting confidence scores carry almost no variation and cannot discriminate between correct and incorrect code. The pattern is slightly less severe on Java, where NLI-based clustering occasionally produces 2--3 clusters, consistent with the narrower but still present performance gap observed in the main results.

Functional equivalence clustering produces a substantially broader distribution of cluster counts across all languages. On Python, cluster counts range from 1 to 11 with meaningful mass across the range, particularly for models with lower accuracy (e.g., GPT-4o-mini), where incorrect solutions are more likely to diverge from each other. On SQL, the distribution is similarly broad, with GPT-4o-mini again showing the widest spread. Java shows an intermediate pattern, with functional equivalence clustering producing more variation than NLI but less than Python or SQL, consistent with the moderate AUROC gains observed for Java in the main text.

These distributions confirm that the performance gap between NLI-based and functional equivalence methods is driven by the inability of NLI models to form meaningful clusters from code, rather than by differences in the summary statistics (entropy, sets confidence) applied to those clusters.

\section{Execution-Based Validation of Equivalence Judgments}
\label{sec:validation}

\subsection{False Equivalence Analysis}
The functional equivalence methods rely on an LLM to judge whether two code snippets are functionally equivalent. To validate the accuracy of these judgments, we compare LLM-assessed equivalence clusters against execution-based outcomes using Gemini-2.5-Flash as the equivalence judge, consistent with the main experiments. Specifically, we check whether responses assigned to the same cluster produce the same results when executed. For Python, we compare actual outputs on test inputs, requiring all cluster members to produce identical results.\footnote{We omit Java from this validation as it uses the same equivalence prompt as Python and the same judge model.} For SQL, we compare pass/fail outcomes against the reference solution, which is a weaker check since two queries can both fail for different reasons. Importantly, we note that this validation can detect false equivalence (the judge incorrectly merges functionally different code into the same cluster) but cannot detect false non-equivalence (the judge incorrectly splits equivalent code into separate clusters), since agreement on a finite set of test inputs does not guarantee equivalence on all possible inputs.

We define \emph{cluster purity} as the proportion of multi-element clusters in which all members produce the same execution results: identical outputs on test inputs for Python, and the same pass/fail outcome for SQL.\footnote{The SQL validation uses pass/fail agreement rather than output comparison because LiveSQLBench's evaluation framework returns only binary correctness against the reference solution; comparing full result sets across sampled queries would have required substantial modifications to the evaluation pipeline. We note that this is a weaker check, since two queries can both fail for different reasons.} Cluster purity is analogous to precision: high purity confirms that the judge rarely merges functionally different code into the same cluster (low false equivalence), but does not measure whether equivalent code is incorrectly split into separate clusters (false non-equivalence). We additionally report \emph{response purity}, the proportion of responses in multi-element clusters that belong to pure clusters, which accounts for variation in cluster size. We evaluate on the highest-accuracy and lowest-accuracy models for each benchmark to bound the analysis across the capability range. For Python, we restrict the analysis to the 444 LeetCode callable-style problems where output comparison is feasible, and exclude instances containing any runtime errors (8 for Gemini-2.5-Pro, 29 for GPT-4o-mini).\footnote{When including clusters with runtime errors and treating matching error types as equivalent, cluster purity drops slightly: 94.7\% (Gemini-2.5-Pro) and 93.5\% (GPT-4o-mini). We report the error-excluded results as the primary metric since runtime errors do not produce meaningful outputs for comparison.} Table~\ref{tab:cluster_purity} summarizes the full results.

\paragraph{Python (LiveCodeBench).} On Gemini-2.5-Pro (highest accuracy), 96.1\% of 560 multi-element clusters are pure. On GPT-4o-mini (lowest accuracy), 96.1\% of 616 multi-element clusters are pure. The consistency across models of different capability levels suggests that the equivalence judge performs reliably regardless of the quality of code being assessed.

\paragraph{SQL (LiveSQLBench).} On Gemini-2.5-Pro (highest accuracy), 96.7\% of 487 multi-element clusters are pure. On GPT-4o-mini (lowest accuracy), 93.1\% of 303 multi-element clusters are pure. The slightly lower purity on GPT-4o-mini likely reflects the greater diversity of incorrect queries generated by the lower-accuracy model.

\begin{table}[htp]
\centering
\tiny
\begin{tabular}{llrrrr}
\toprule
 & & \multicolumn{2}{c}{Cluster-level} & \multicolumn{2}{c}{Response-level} \\
\cmidrule(lr){3-4} \cmidrule(lr){5-6}
Benchmark & Model & Clusters & Purity & Responses & Purity \\
\midrule
\multirow{2}{*}{Python} & Gem-2.5-Pro & 560 & 96.1\% & 4,586 & 97.1\% \\
 & GPT-4o-mini & 616 & 96.1\% & 3,396 & 97.2\% \\
\midrule
\multirow{2}{*}{SQL} & Gem-2.5-Pro & 487 & 96.7\% & 1,731 & 96.7\% \\
 & GPT-4o-mini & 303 & 93.1\% & 1,032 & 92.2\% \\
\bottomrule
\end{tabular}
\caption{Cluster purity of LLM-based equivalence judgments validated against execution. Cluster-level purity is the proportion of multi-element clusters where all members produce the same execution results. Response-level purity is the proportion of responses in multi-element clusters that belong to pure clusters.}
\label{tab:cluster_purity}
\end{table}

\paragraph{Summary.} Across both benchmarks and all four model evaluations, cluster purity exceeds 93\%, indicating that the LLM equivalence judge rarely merges functionally different code into the same cluster. The validation is one-directional: high purity confirms low false equivalence rates, but does not measure false non-equivalence, since agreement on finite test inputs does not guarantee equivalence on all possible inputs. Nonetheless, false equivalence is the more consequential error for uncertainty quantification, as it would inflate confidence scores for incorrect code.

\subsection{False Non-Equivalence Analysis}
The cluster purity analysis above measures false equivalence (the judge incorrectly merging functionally different code), but does not measure the complementary failure mode: false non-equivalence (the judge incorrectly splitting functionally equivalent code into separate clusters). To quantify this, we compute a response-level misclassification rate using execution-based ground truth.

For Python, where we have full execution outputs, we group responses by identical output on the test suite. For each such group, we identify the largest judge cluster containing members of that group and count the remaining responses as misclustered. For SQL, where we only have binary pass/fail outcomes against the reference solution, we take all correct responses (those passing all test cases) as a single equivalence group and similarly measure how many are split away from the majority cluster. In both cases, the FNE rate is the fraction of equivalent (or correct) responses that are misclustered.

These rates are approximate upper bounds on true false non-equivalence: test suites are finite and may not cover edge cases where two responses actually differ in behavior, meaning the judge may correctly distinguish responses that appear equivalent under limited test inputs. Table~\ref{tab:fne_analysis} summarizes the results.

\begin{table}[htp]
\centering
\scriptsize
\begin{tabular}{llcc}
\toprule
Benchmark & Model & FNE Rate & Ground Truth \\
\midrule
\multirow{2}{*}{Python} & Gem-2.5-Pro & 10.3\% & Output match \\
 & GPT-4o-mini & 17.9\% & Output match \\
\midrule
\multirow{2}{*}{SQL} & Gem-2.5-Pro & 45.1\% & Pass/fail (upper bound) \\
 & GPT-4o-mini & 42.4\% & Pass/fail (upper bound) \\
\bottomrule
\end{tabular}
\caption{False non-equivalence (FNE) analysis. Response-level misclassification rate: fraction of execution-equivalent (Python) or correct (SQL) responses assigned to a minority cluster. SQL rates are upper bounds, as binary pass/fail does not guarantee functional equivalence.}
\label{tab:fne_analysis}
\end{table}

For Python, the FNE misclassification rates are moderate (10.3--17.9\%), indicating that the judge occasionally splits functionally identical code into separate clusters but does so for a relatively small fraction of responses. For SQL, the upper-bound rates are higher (42.4--45.1\%), reflecting the greater syntactic diversity of equivalent SQL formulations. Importantly, false non-equivalence inflates entropy conservatively, causing the system to overestimate rather than underestimate uncertainty. As shown in Tables~\ref{tab:auroc_sql} and~\ref{tab:auroc_python}, AUROC scores remain strong across all settings despite this inflation, confirming that the relative ranking of uncertainty is preserved. Domain-specific equivalence strategies, such as SQL-aware normalization prior to judging, represent a promising direction for reducing false non-equivalence in future work.

\section{Computational Cost Analysis}
\label{sec:cost}

\begin{table*}[h]
\centering
\small
\begin{tabular}{lccc}
\toprule
\textbf{Method Family} & \textbf{Generations (Orig.\ LLM)} & \textbf{Equiv.\ Judge Calls} & \textbf{Logprobs} \\
\midrule
Token-probability        & 1     & 0                                            & Yes \\
Reflexive                & 1 + 1\textsuperscript{\dag}  & 0                  & P(True): Yes; VC: No \\
Similarity-based         & $m+1$ & 0                                            & No \\
Functional equivalence   & $m+1$ & FER: $m$; Clustering: up to $\binom{m+1}{2}$   & No \\
\bottomrule
\end{tabular}
\caption{Computational cost by method family. \textit{Generations (Orig.\ LLM)} denotes calls to the original language model. \textit{Equiv.\ Judge Calls} denotes calls to the equivalence judge model. \textit{Logprobs} indicates whether access to token-level log-probabilities is required. \textsuperscript{\dag}Reflexive methods use one additional call to the original LLM for self-evaluation, not the equivalence judge.}
\label{tab:cost-complexity}
\end{table*}

We summarize the computational requirements of each method family in terms of calls to the original LLM, calls to the equivalence judge, and whether access to log-probabilities is required.
Token-probability methods are the cheapest, requiring only a single greedy generation with log-probabilities enabled. Reflexive methods add one self-evaluation call to the original LLM. P(True) extracts the token probability assigned to ``True,'' while verbalized confidence prompts the model to express its confidence on a Likert scale. Neither requires an equivalence judge. Similarity-based methods (cosine similarity, CodeBLEU) require $m + 1$ generations but no additional LLM calls, as similarity computation is handled locally by the embedding model or CodeBLEU. 

Functional equivalence methods require $m + 1$ generations from the original LLM plus additional calls to the equivalence judge. Specifically, functional equivalence rate requires $m$ pairwise comparisons between the original response and each sample, while the clustering-based methods (functional entropy, functional sets confidence) require up to $\binom{m+1}{2}$ pairwise comparisons among all sampled responses. In practice, the clustering algorithm terminates early once cluster assignments are determined, so the number of comparisons is typically well below the $\binom{m+1}{2}$ upper bound. All equivalence calls are independent, and our implementation executes them as asynchronous LLM calls, so wall-clock latency scales less than linearly with the number of comparisons.

\paragraph{Cost structure of equivalence assessment.} Each equivalence call consists of the prompt template (Appendix~\ref{sec:prompts}) plus two code snippets, with the judge returning a single classification label (``Equivalent'' or ``Not Equivalent''). Output token costs are therefore negligible, and the cost of equivalence assessment is dominated by input tokens, which scale with the length of generated solutions and vary by benchmark. At $m = 5$ (the practical recommendation from Appendix~\ref{sec:ablations}), functional equivalence rate requires 5 judge calls per prompt, while clustering-based methods require up to 15. At $m = 10$, these increase to 10 and up to 55, respectively. In both cases, individual equivalence calls are substantially cheaper than generation calls, since they produce only the predicted class rather than a full code response.


\paragraph{Dollar cost analysis.} To ground the above complexity analysis in concrete terms, we report the average per-prompt dollar cost of generation and equivalence scoring across all model--language combinations using Gemini-2.5-Flash as the equivalence judge (\$0.30 per 1M input tokens, \$2.50 per 1M output tokens). Table~\ref{tab:model_costs} reports three cost components: (1) the generation cost for a single response from the primary LLM, (2) the sampled generation cost for $m=10$ additional responses from the primary LLM, and (3) the equivalence scoring cost. The scoring cost column reports a range corresponding to the best case ($m$ judge calls, when all samples are assigned to one cluster via early transitivity) and the worst case ($\binom{m+1}{2}$ pairwise calls, when all samples are in separate clusters).

Across all settings, equivalence scoring costs are modest and generally comparable to or smaller than the sampled generation cost. For example, for GPT-4o on Python, sampled generation costs \$0.036 per prompt while scoring adds \$0.002--\$0.010. The most expensive generation configuration is GPT-4o on SQL (\$0.406 per prompt for sampled generation), where scoring adds only \$0.003--\$0.016. This indicates that for practitioners already committing to multi-sample generation for sampling-based UQ, the marginal cost of functional equivalence scoring is reasonable relative to the generation cost that all sampling-based methods share.

\begin{table}[htbp]
\centering
\tiny
\begin{tabular}{llrrr}
\toprule
\multirow{2}{*}{Lang.} & \multirow{2}{*}{Model} & Generation & Sampled Gen. & Scoring Cost \\
 &  & (\$) & (\$) & (\$ Lower--Upper) \\
\midrule
\multirow{5}{*}{\rotatebox[origin=c]{90}{Java}} & 25-flash & 4.7e-04 & 0.004 & 0.002--0.008 \\
 & 25-flash-lite & 6.9e-05 & 5.0e-04 & 0.001--0.007 \\
 & 25-pro & 0.001 & 0.010 & 0.001--0.006 \\
 & 4o & 9.9e-04 & 0.009 & 0.001--0.006 \\
 & 4o-mini & 5.9e-05 & 5.7e-04 & 0.001--0.006 \\
\midrule
\multirow{5}{*}{\rotatebox[origin=c]{90}{Python}} & 25-flash & 0.002 & 0.017 & 0.004--0.025 \\
 & 25-flash-lite & 0.001 & 0.005 & 0.009--0.050 \\
 & 25-pro & 0.009 & 0.073 & 0.005--0.025 \\
 & 4o & 0.004 & 0.036 & 0.002--0.010 \\
 & 4o-mini & 1.9e-04 & 0.002 & 0.002--0.009 \\
\midrule
\multirow{5}{*}{\rotatebox[origin=c]{90}{SQL}} & 25-flash & 0.006 & 0.054 & 0.003--0.017 \\
 & 25-flash-lite & 0.002 & 0.016 & 0.003--0.017 \\
 & 25-pro & 0.026 & 0.257 & 0.003--0.016 \\
 & 4o & 0.041 & 0.406 & 0.003--0.016 \\
 & 4o-mini & 0.002 & 0.024 & 0.002--0.013 \\
\bottomrule
\end{tabular}
\caption{Average dollar cost per prompt (\$USD) for generation and equivalence scoring by language and model, using Gemini-2.5-Flash as the equivalence judge. \textit{Generation} is the cost of a single original response. \textit{Sampled Generation} is the cost of $m{=}10$ additional sampled responses. \textit{Scoring} reports the range from best-case ($m$ judge calls) to worst-case ($\binom{m+1}{2}$ judge calls).}
\label{tab:model_costs}
\end{table}

\section{Ablations}
\label{sec:ablations}
\subsection{Number of Sampled Responses}
\label{sec:sampled_ablation}
\begin{figure*}[tb]
    \centering
    \includegraphics[width=\textwidth]{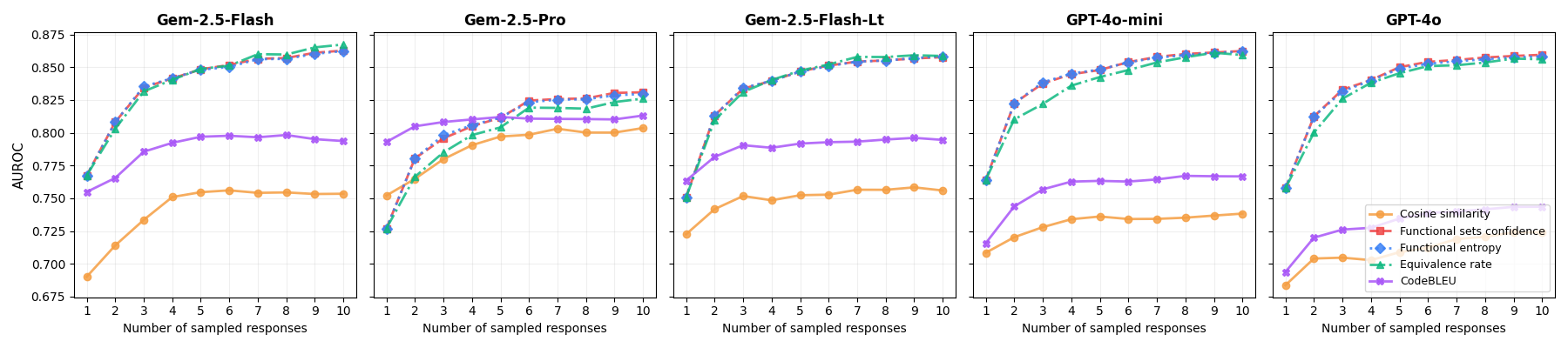}
    \caption{AUROC as a function of the number of sampled responses $m$ on LiveCodeBench (Python). Performance increases with $m$ but exhibits diminishing returns, with most gains realized by $m=5$.}
    \label{fig:m_ablation_python}
\end{figure*}

\begin{figure*}[tb]
    \centering
    \includegraphics[width=\textwidth]{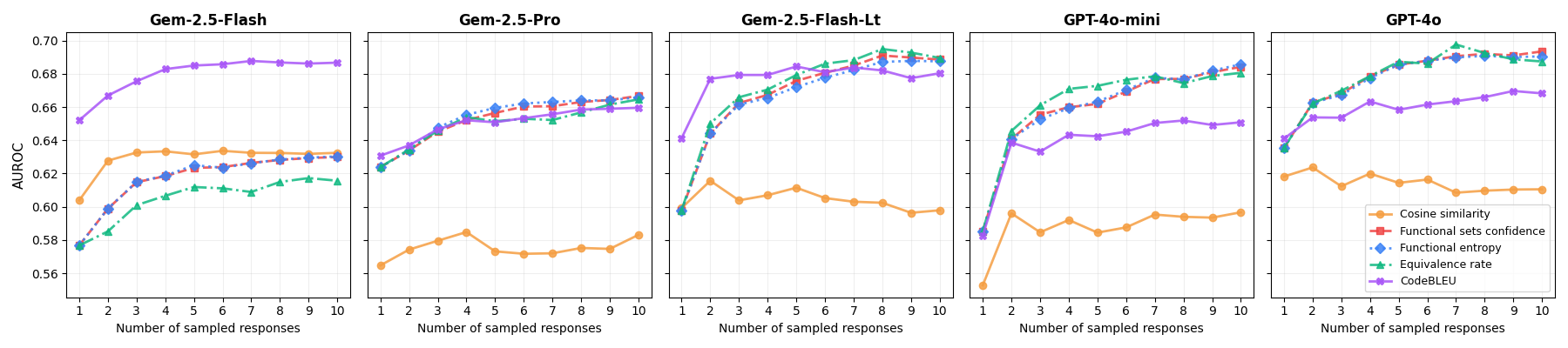}
    \caption{AUROC as a function of the number of sampled responses $m$ on MultiPL-E (Java). Performance increases with $m$ but exhibits diminishing returns, with most gains realized by $m=5$.}
    \label{fig:m_ablation_java}
\end{figure*}

\begin{figure*}[tb]
    \centering
    \includegraphics[width=\textwidth]{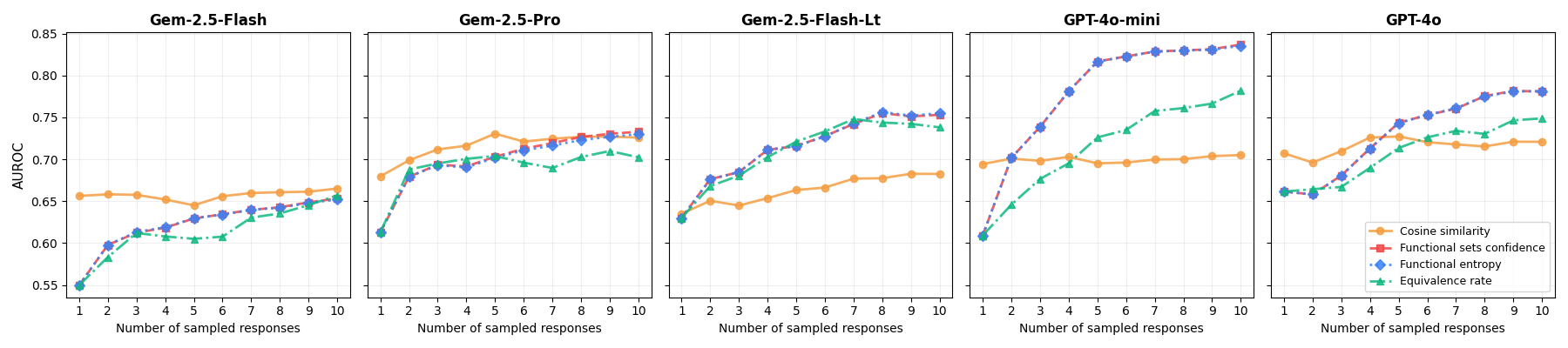}
    \caption{AUROC as a function of the number of sampled responses $m$ on LiveSQLBench (SQL). The pattern is similar to Python, with more variance at low $m$ due to the smaller benchmark size.}
    \label{fig:m_ablation_sql}
\end{figure*}

We evaluate the sensitivity of sampling-based methods to the number of sampled responses $m \in \{1, \ldots, 10\}$. Figures~\ref{fig:m_ablation_python},~\ref{fig:m_ablation_java}, and~\ref{fig:m_ablation_sql} present results for Python, Java, and SQL, respectively.

Across all three languages, AUROC increases with $m$ but exhibits diminishing returns. On Python, functional equivalence methods improve steeply from $m=1$ to $m=4$, after which gains flatten considerably. By $m=5$, most models achieve within 0.01--0.02 AUROC of their $m=10$ performance. Functional entropy and functional sets confidence follow nearly identical trajectories, consistent with both methods relying on the same underlying cluster structure. Equivalence rate shows a similar pattern but starts slightly higher at low $m$, as it does not require multiple clusters to produce a useful signal. Cosine similarity is notably less sensitive to $m$, with performance plateauing by $m=3$ on most models, likely because pairwise similarity scores are informative even with few samples.

On Java, the diminishing returns pattern holds but with two notable differences. First, CodeBLEU consistency is competitive with or exceeds functional equivalence methods at all values of $m$, particularly on Gemini-2.5-Flash where it achieves the highest AUROC throughout. This contrasts with Python and SQL, where functional equivalence methods dominate at higher $m$. Second, equivalence rate lags behind functional entropy and functional sets confidence more substantially than on other languages, suggesting that the binary equivalence signal is less informative for Java code where functional differences may be more subtle. Cosine similarity again plateaus early but at a lower level than on Python.

On SQL, the overall pattern is similar but with more variance at low $m$, reflecting the smaller benchmark size (270 problems). Functional entropy and functional sets confidence again show the steepest improvement, with most gains realized by $m=5$--$6$. Cosine similarity observes generally smaller gains from additional samples relative to Python and Java. Equivalence rate improves more gradually on SQL and does not fully converge by $m=10$ for some models (e.g., GPT-4o-mini), indicating that pairwise equivalence judgments for SQL may benefit from larger sample sizes.

Overall, these results suggest that $m=5$ offers a practical balance between performance and computational cost for most settings.

\subsection{Sensitivity to Equivalence Judge}
\label{sec:judge-sensitivity}

The functional equivalence methods introduced in Section~\ref{sec:consistency} rely on an LLM to assess pairwise functional equivalence between code responses. All main experiments use Gemini-2.5-Flash as the equivalence judge. To evaluate how sensitive the methods are to judge choice, we replace the judge with two alternatives: Gemini-2.5-Flash-Lite, a smaller model from the same provider, and GPT-4o-mini, a model from a different provider (OpenAI). Following Appendix~\ref{sec:validation}, we evaluate on Python (LiveCodeBench) and SQL (LiveSQLBench) using the highest- and lowest-accuracy generators for each benchmark: Gemini-2.5-Pro and GPT-4o-mini.

Table~\ref{tab:judge-sensitivity} presents AUROC results for the three functional equivalence scorers across all three judges, with the Gemini-2.5-Flash results from the main experiments included for reference.

\begin{table}[t]
\centering
\tiny
\begin{tabular}{ll ccc}
\toprule
\textbf{Original LLM} & \textbf{Scorer} & \textbf{Gem-Fl.} & \textbf{Gem-Fl.-Lt} & \textbf{GPT4o-m} \\
\midrule
\multicolumn{5}{l}{\textit{Python (LiveCodeBench)}} \\
\midrule
\multirow{3}{*}{Gem-Pro}
 & Func.\ Entropy   & 0.829 & 0.834 & 0.829 \\
 & Func.\ Sets      & 0.826 & 0.836 & 0.831 \\
 & Equiv.\ Rate     & 0.826 & 0.829 & 0.814 \\
\midrule
\multirow{3}{*}{GPT4o-m}
 & Func.\ Entropy   & 0.859 & 0.852 & 0.836 \\
 & Func.\ Sets      & 0.859 & 0.855 & 0.837 \\
 & Equiv.\ Rate     & 0.856 & 0.842 & 0.824 \\
\midrule
\multicolumn{5}{l}{\textit{SQL (LiveSQLBench)}} \\
\midrule
\multirow{3}{*}{Gem-Pro}
 & Func.\ Entropy   & 0.745 & 0.752 & 0.711 \\
 & Func.\ Sets      & 0.750 & 0.752 & 0.714 \\
 & Equiv.\ Rate     & 0.702 & 0.700 & 0.637 \\
\midrule
\multirow{3}{*}{GPT4o-m}
 & Func.\ Entropy   & 0.840 & 0.828 & 0.825 \\
 & Func.\ Sets      & 0.842 & 0.828 & 0.826 \\
 & Equiv.\ Rate     & 0.782 & 0.791 & 0.756 \\
\bottomrule
\end{tabular}
\caption{AUROC of functional equivalence methods across three equivalence judges. Gem-Fl. denotes the primary judge (Gemini-2.5-Flash) used in all main experiments. Gemini-2.5-Flash-Lite and GPT-4o-mini are alternative judges of varying capability.}
\label{tab:judge-sensitivity}
\end{table}

Gemini-2.5-Flash-Lite produces AUROC scores within 0.01 of the primary judge in most configurations, and in some cases marginally exceeds it (e.g., functional entropy on Python with Gemini-2.5-Pro responses: 0.834 vs.\ 0.829). This suggests that the functional equivalence approach is not sensitive to moderate reductions in judge capability within the same model family.

GPT-4o-mini shows more notable degradation as a judge. On Python, functional entropy shows no degradation on Gemini-2.5-Pro responses but drops by 0.023 on GPT-4o-mini responses relative to the primary judge. On SQL, the drop is larger: 0.034 on Gemini-2.5-Pro responses and 0.015 on GPT-4o-mini responses. Equivalence rate is the most affected scorer, dropping by 0.065 on SQL with Gemini-2.5-Pro responses, likely because it relies on individual pairwise judgments without the smoothing effect of cluster-level aggregation. Functional entropy and functional sets confidence degrade more gracefully, consistent with their reliance on distributional properties of the full cluster structure rather than individual binary comparisons.

Despite the degradation with GPT-4o-mini, functional equivalence methods remain competitive with or exceed the strongest baselines in most configurations, indicating that the approach provides value even with a weaker judge. These results suggest that judge capability affects absolute performance but does not alter the overall conclusion that LLM-based functional equivalence assessment improves upon both NLI-based and token-probability methods for code correctness prediction.

\subsection{Sensitivity to Sampled Response Ordering}
\label{app:ordering}

The greedy clustering procedure used by Functional Sets Confidence and Functional Negentropy is order-dependent: the sequence in which responses are compared can affect cluster assignments. To assess the impact of this ordering sensitivity on downstream UQ metrics, we conduct a permutation analysis. For both Python (LiveCodeBench) and SQL (LiveSQLBench), we sample 100 problems for both the strongest model (Gemini-2.5-Pro) and weakest model (GPT-4o-mini), and recompute all three functional equivalence scorers 10 times with random permutations of the $m=10$ sampled responses. We note that Functional Equivalence Rate is permutation-invariant by construction, as it aggregates all pairwise judgments without clustering. Any variation observed for this scorer therefore reflects LLM judge stochasticity rather than ordering effects.

Table~\ref{tab:ordering_sensitivity} summarizes the results. Across all 12 scorer--dataset--model combinations, AUROC ranges are narrow, and pairwise Pearson correlations between runs exceed 0.82 in all settings and 0.94 in 10 of 12 settings. ECE variation is similarly small. We note that this analysis uses subsamples of 100 problems, and metric variability is expected to decrease on the full dataset due to standard sampling effects.

\begin{table}[htp]
\centering
\tiny
\begin{tabular}{llccc}
\toprule
Scorer & Setting & AUROC Range & ECE Range & Min $r$ \\
\midrule
\multirow{4}{*}{Func. Sets}
& SQL / Gem-2.5-Pro & 0.71--0.75 & 0.09--0.13 & 0.95 \\
& SQL / 4o-mini & 0.80--0.83 & 0.12--0.17 & 0.96 \\
& Py / Gem-2.5-Pro & 0.87--0.94 & 0.02--0.06 & 0.83 \\
& Py / 4o-mini & 0.86--0.88 & 0.07--0.11 & 0.98 \\
\midrule
\multirow{4}{*}{Func. Entr.}
& SQL / Gem-2.5-Pro & 0.71--0.74 & 0.15--0.18 & 0.95 \\
& SQL / 4o-mini & 0.80--0.83 & 0.08--0.14 & 0.96 \\
& Py / Gem-2.5-Pro & 0.89--0.94 & 0.05--0.07 & 0.88 \\
& Py / 4o-mini & 0.87--0.89 & 0.04--0.08 & 0.98 \\
\midrule
\multirow{4}{*}{Func. Equiv.}
& SQL / Gem-2.5-Pro & 0.64--0.69 & 0.24--0.27 & 0.94 \\
& SQL / 4o-mini & 0.74--0.80 & 0.08--0.14 & 0.96 \\
& Py / Gem-2.5-Pro & 0.94--0.96 & 0.09--0.11 & 0.91 \\
& Py / 4o-mini & 0.89--0.91 & 0.10--0.13 & 0.98 \\
\bottomrule
\end{tabular}
\caption{Ordering sensitivity of functional equivalence scorers. For each setting, we report the AUROC and ECE range (min--max) across 10 reshuffled runs and the minimum pairwise Pearson correlation between runs.}
\label{tab:ordering_sensitivity}
\end{table}

\section{Prompt Templates}
\label{sec:prompts}
The complete equivalence prompt used in our Python and Java experiments is presented in Table~\ref{tab:python_java_prompt}, while the prompt for SQL experiments is presented in Table \ref{tab:sql_prompt}.
\begin{table*}[tb]
\centering
\begin{tcolorbox}[colback=blue!5!white, colframe=blue!50!black, title = {\textsc{Python/Java Code Equivalence}}]
\ttfamily

You are a [LANGUAGE] code equivalence judge. \\

Definition:\\
Two [LANGUAGE] code blocks are considered functionally equivalent if they would produce the same outputs for the same inputs.\\
\\
Consider equivalent:\\
- Different implementations or algorithms that achieve the same result\\
- Refactored or restructured code with the same behavior\\
- Minor variations in edge case handling\\
\\
Consider NOT equivalent:\\
- Code that would produce different outputs for the same inputs\\
- Code where one is incomplete or missing functionality present in the other\\
\\
Decision rule:\\
- If both code snippets would generally produce the same results → output exactly: "Equivalent"\\
- If the code snippets would produce different outputs → output exactly: "Not Equivalent"\\
\\
Output format:\\
Output EXACTLY one of: "Equivalent" OR "Not Equivalent".\\
Do not add explanations, reasoning, punctuation, or extra text.\\

\end{tcolorbox}
\caption{Prompt for Python/Java Code Equivalence.}
\label{tab:python_java_prompt}
\end{table*}

\begin{table*}[tb]
\centering
\begin{tcolorbox}[colback=blue!5!white, colframe=blue!50!black, title = {\textsc{SQL Equivalence}}]
\ttfamily

You are a SQLite query equivalence judge.
\\
\\
Definition:\\
Two SQLite queries are considered semantically equivalent only if, when executed against the same database state, they produce exactly the same result set.
\\
\\
Same result set means:\\
- The same rows (treating rows as unordered sets, unless ORDER BY is specified in both queries)\\
- The same column values in each row\\
- The same column order\\
\\
Ignore:\\
- Purely syntactic differences (formatting, whitespace, capitalization of keywords)\\
- Use of aliases that do not affect the result set\\
- Equivalent expressions (e.g., `WHERE a = 1 AND b = 2` vs. `WHERE b = 2 AND a = 1`)\\
- Different join syntax with equivalent semantics (e.g., implicit vs. explicit JOIN)
- Use of parentheses that do not change query semantics\\
- Comments
\\
\\
Do NOT ignore:\\
- Different row ordering if either query specifies ORDER BY\\
- NULL handling differences that affect results\\
- DISTINCT vs. non-DISTINCT if it changes output rows\\
- Column order differences in the SELECT clause
\\
\\
Decision rule:\\
- If both queries would return the same result set on any valid database state → output exactly: "Equivalent"\\
- If any valid database state exists where the queries would return different results → output exactly: "Not Equivalent"
\\
\\
Output format: \\
Output EXACTLY one of: "Equivalent" OR "Not Equivalent".\\
Do not add explanations, reasoning, punctuation, or extra text.

\end{tcolorbox}
\caption{Prompt for SQLite Equivalence.}
\label{tab:sql_prompt}
\end{table*}

\end{document}